\documentclass[10pt,twocolumn,letterpaper]{article}

\usepackage{iccv}
\usepackage{amssymb}
\usepackage{pbox}
\usepackage{times}
\usepackage{epsfig}
\usepackage{graphicx}
\usepackage{amsmath}
\usepackage{amsfonts}
\usepackage{multicol}
\usepackage{float}
\usepackage{multirow}
\usepackage{graphics}
\usepackage{caption}
\usepackage{pifont}
\usepackage{gensymb}
\usepackage{xcolor}
\usepackage{makecell}
\usepackage[ruled,vlined]{algorithm2e}
\usepackage{algorithmic}
\usepackage{mathtools}
\usepackage{subfigure}
\usepackage{booktabs}
\usepackage{xspace}
\usepackage{enumitem}
\usepackage{pdfpages}
\usepackage{wrapfig}
\usepackage{siunitx}
\NewDocumentCommand{\codeword}{v}{%
\texttt{\textcolor{black}{#1}}%
}

\newcommand{\xmark}{\textcolor{red}{\ding{55}}}%
\newcommand{\cmark}{\textcolor{green}{\checkmark}}%

\newcommand{\name}{\textit{CSP}\xspace}

\usepackage[pagebackref=true,breaklinks=true,letterpaper=true,colorlinks,bookmarks=false]{hyperref}
\iccvfinalcopy 

\def\iccvPaperID{9228} 

\begin{document}

\title{Deep Implicit Surface Point Prediction Networks}



\author{
	Rahul Venkatesh\textsuperscript{1} \ \ \ 
	Tejan Karmali\textsuperscript{2} \ \ \ 
	Sarthak Sharma\textsuperscript{3} \ \ \ 
	Aurobrata Ghosh\textsuperscript{3} \\
	R. Venkatesh Babu\textsuperscript{2} \ \ \ \ \ \ 
	L{\'a}szl{\'o} A. Jeni\textsuperscript{1} \ \ \ \ \ \ 
	Maneesh Singh\textsuperscript{3} \\

	\small \textsuperscript{1}Carnegie Mellon University, Pittsburgh, PA, USA \ \ \ \ \ \textsuperscript{2}Indian Institute of Science, Bengaluru, India \\ \small \textsuperscript{3}Verisk Analytics, Jersey City, NJ, USA
}



\twocolumn[{%
\maketitle

    \centering
    \includegraphics[width=1.0\textwidth]{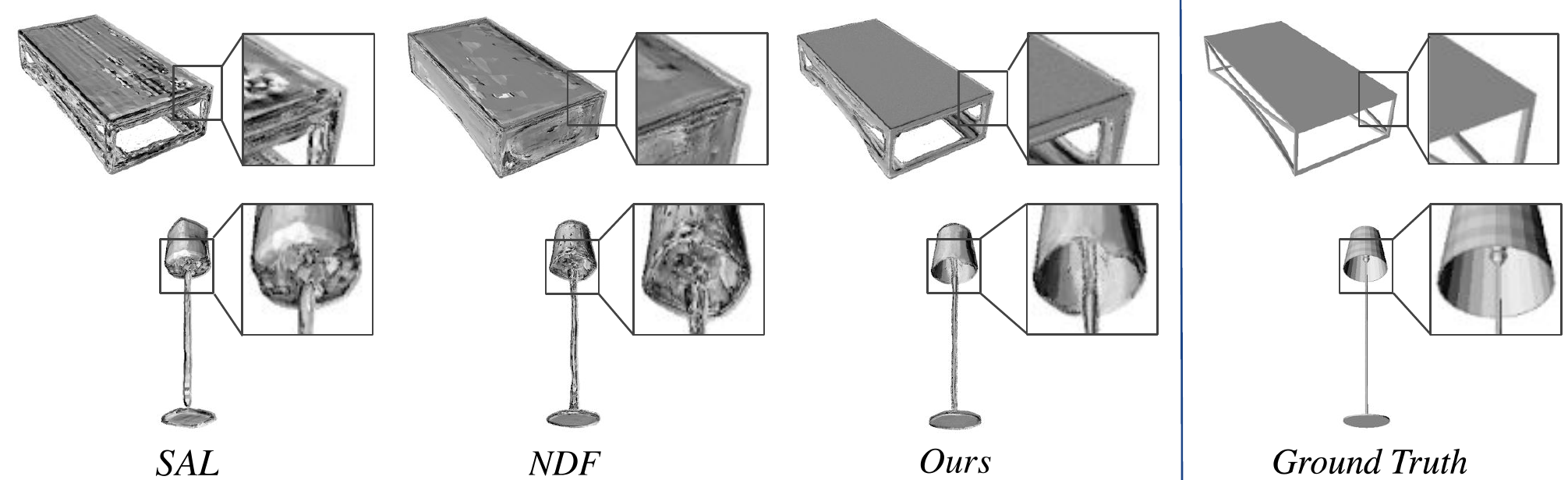}
    \captionof{figure}{Our novel implicit shape representation can model complex surfaces with high-fidelity. \textit{\textbf{Row 1:}} Recovering visually pleasing surfaces in comparison to prior state-of-the-art SAL~\cite{sal-2020} and NDF~\cite{NEURIPS2020_f69e505b}. \textit{\textbf{Row 2:}} Results on a representative open shape, where we correctly model the shape, as opposed to SAL~\cite{sal-2020}, which closes up regions that are meant to be open.}
    \label{fig:shape_recon}
    \vspace*{5mm}
}]
\ificcvfinal\thispagestyle{empty}\fi
\setcounter{figure}{1}

\begin{abstract}
\vspace*{-0.2cm}    
    \noindent Deep neural representations of 3D shapes as implicit functions have been shown to produce high fidelity models surpassing the resolution-memory trade-off faced by the explicit representations using meshes and point clouds. However, most such approaches focus on representing closed shapes. Unsigned distance function (UDF) based approaches have been proposed recently as a promising alternative to represent both open and closed shapes. However, since the gradients of UDFs vanish on the surface, it is challenging to estimate local (differential) geometric properties like the normals and tangent planes which are needed for many downstream applications in vision and graphics. There are additional challenges in computing these properties efficiently with a low-memory footprint. This paper presents a novel approach that models such surfaces using a new class of implicit representations called the closest surface-point (\name) representation. We show that \name allows us to represent complex surfaces of any topology (open or closed) with high fidelity. It also allows for accurate and efficient computation of local geometric properties. We further demonstrate that it leads to efficient implementation of downstream algorithms like sphere-tracing for rendering the 3D surface as well as to create explicit mesh-based representations. Extensive experimental evaluation on the ShapeNet dataset validate the above contributions with results surpassing the state-of- the-art.  

\vspace*{-0.5cm}
\end{abstract}

\section{Introduction}

\label{sec:intro}
  High fidelity representation and rendering of potentially open 3D surfaces with complex topology from raw sensor data (images, point clouds) finds application in vision, graphics and animation industry~\cite{survey3D}. Therefore, in recent years deep learning based methods for 3D reconstruction of objects have garnered significant interest~\cite{qi2017pointnet, deepsdf, occupancy_network}. 
 
\par

Explicit 3D shape representations such as point clouds, voxels, triangles or quad meshes pose challenges in reconstructing surfaces with arbitrary topology ~\cite{pan2019deep}. Moreover, the ability to capture details of such representations are limited by predefined structure (like number of vertices for meshes) or memory and computational footprint (for voxels and point clouds). Several implicit shape representations using deep neural networks have been proposed~\cite{deepsdf,occupancy_network, cvxnet-2020, chen2020bsp, sal-2020, NEURIPS2020_f69e505b} to alleviate these shortcomings. 

Recent approaches use a distance function as the implicit representation. For example, DeepSDF~\cite{deepsdf} use a Signed Distance Function (SDF) as the implicit representation  where the sign represents the inside/ outside of the surface being modeled. Not only does this limit DeepSDF to modeling closed surfaces, the ground truth needs to be watertight (\textit{closed}) as well. Since most 3D shape datasets~\cite{chang2015shapenet} have non-watertight (\textit{open}) shapes, preprocessing is needed to artificially close such shapes and make them watertight~\cite{occupancy_network} - a process which is known to result in a loss of fidelity~\cite{huang2018robust}. To overcome this problem, methods such as SAL~\cite{sal-2020} seek to learn surface representations directly from raw unoriented point clouds. However, such methods also make an assumption that the underlying surface represented by the point cloud is closed, leading to learnt representations necessarily describing closed shapes~\cite{atzmon2021sald}. 

NDF~\cite{NEURIPS2020_f69e505b} overcomes this limitation by using an unsigned distance function (\textit{UDF}) based implicit representation and achieves state-of-the-art performance on 3D shape representation learnt directly from the unprocessed ShapeNet dataset. However, \textit{UDFs} have a fundamental limitation. Since the gradient of the \textit{UDF} vanishes on the surface, direct estimation of local, differential geometric properties like the tangent plane and the surface normal becomes noisy and loses fidelity. This results in a loss of performance on downstream tasks like rendering and surface reconstruction~\cite{kazhdan2006poisson} as well as those like  registration~\cite{pomerleau2015review} and segmentation~\cite{grilli2017review} where normal estimates play a vital role.

An additional issue is that for the above methods, the estimation of differential geometric properties needs a backward pass leading to increased memory footprint and time complexity which becomes a challenge for applications which require fast rendering on devices with limited memory (e.g. tiled rendering on hand-held devices~\cite{correa2003out}), or for robotics tasks such as real-time path planning where fast normal estimates in 3D space play an essential role~\cite{robot_duhautbout2019distributed, robot_oleynikova2016signed}.

To address these challenges, we introduce a novel implicit shape representation called the \textit{Closest Surface-Point} (\textit{\name}) function, which for a given query point returns the closest point on the surface. We demonstrate that \textit{\name} can model open and closed shapes of arbitrary topology, and in contrast to NDF, allows for the easy computation of differential geometry properties like the tangent plane and the surface normal. Moreover, as opposed to existing implicit representations and demonstrated later, it can efficiently recover normal information with a forward pass. A comparative summary of the properties discussed above for \name and the most related art is presented in Table~\ref{table:rep power}. We also present a panel of illustrative results in Fig.~\ref{fig:shape_recon} which clearly demonstrates the higher fidelity with which complex surfaces are represented by \name when compared with \textit{SAL} and \textit{NDF}. 

Finally, we show that due to the above benefits, \name is not only a potential method of choice for learning high fidelity 3D representations of complex topologies (open as well as closed) from the raw data, but also for many downstream applications. For this, we present (a) a fast and memory efficient rendering algorithm using an adaptation of sphere-tracing for \textit{\name}\textit{s} that leverages the accurate surface normal estimates that \textit{\name} provides, and, (b) since it's often required to extract an explicit surface representation~\cite{akkouche2001adaptive}, we present a coarse to fine meshing algorithm for \textit{\name}\textit{s}, that can recover high-fidelity meshes faster than prior methods~\cite{NEURIPS2020_f69e505b}. 
To summarize, our contributions are: 
\begin{itemize}[noitemsep,nolistsep]
    \item \textit{\name}: A high fidelity representation capable of modelling both open and closed shapes, allowing for efficient estimation of differential geometric properties of the surface (Sec.~\ref{sec:shape_rep}) - an advancement over NDF.
    \item Normal estimation with a forward pass that significantly accelerates speed and memory efficiency of rendering (Sec.~\ref{sec:normals}).
    \item A novel sphere tracing (\textit{ST}) algorithm using \name to obtain more accurate renderings over vanilla \textit{ST} for \textit{UDF}. (Sec.~\ref{sec:spheretracing}).
    \item A faster multi-resolution surface extraction technique (Sec.~\ref{sec:meshing}) to extract meshes from \textit{\name}, achieving better speed and quality than existing techniques~\cite{NEURIPS2020_f69e505b}.
\end{itemize}

\begin{table}[t!]
\resizebox{\columnwidth}{!}{
\begin{tabular}{c|c|c|c|c|c}
\hline
    \multirow{3}{*}{Method} & \multirow{3}{*}{\makecell{ Learning \\ from\\ Triangle Soups}} & \multicolumn{3}{c|}{Representation Power} & \multirow{3}{*}{\makecell{ Single pass \\ normal \\ estimation}}\\
    \cline{3-5}
     &  & \makecell{Open\\Shapes} & \makecell{Complex \\ Topology} & \makecell{High \\ Fidelity} &\\
    \hline
     DeepSDF~\cite{deepsdf}& \xmark& \xmark & \cmark & \cmark  & \xmark\\
     SAL~\cite{sal-2020} & \cmark & \xmark & \cmark & \xmark  & \xmark \\
     NDF~\cite{NEURIPS2020_f69e505b} & \cmark & \cmark & \cmark & \xmark  & \xmark \\
     \hline
     \name \textit{(Ours)} & \cmark & \cmark & \cmark & \cmark  & \cmark  \\
\hline
\end{tabular}
}
\vspace{0.5mm}
\caption{Comparison between \name and closely related arts.}
\vspace*{-2mm}
\label{table:rep power}
\end{table}

\section{Related Work}
\label{sec:related_wk}

\begin{figure*}[t!]
  \centering
  \subfigure[Overview of the proposed system ]{\includegraphics[width=0.74\linewidth]{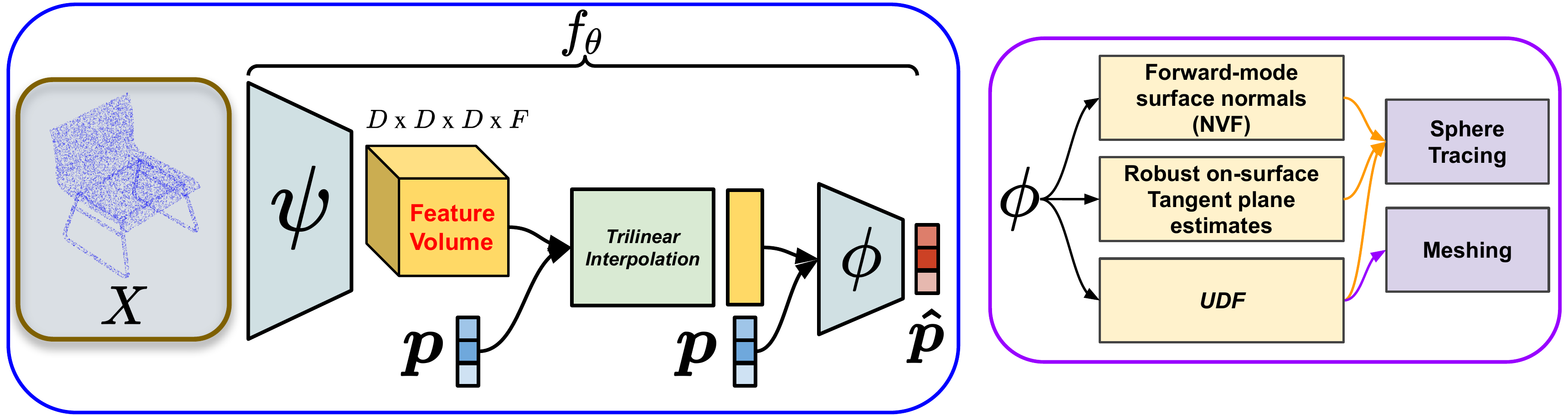}}\quad
  \subfigure[Estimation of local geometric properties]{\includegraphics[width=0.23\linewidth]{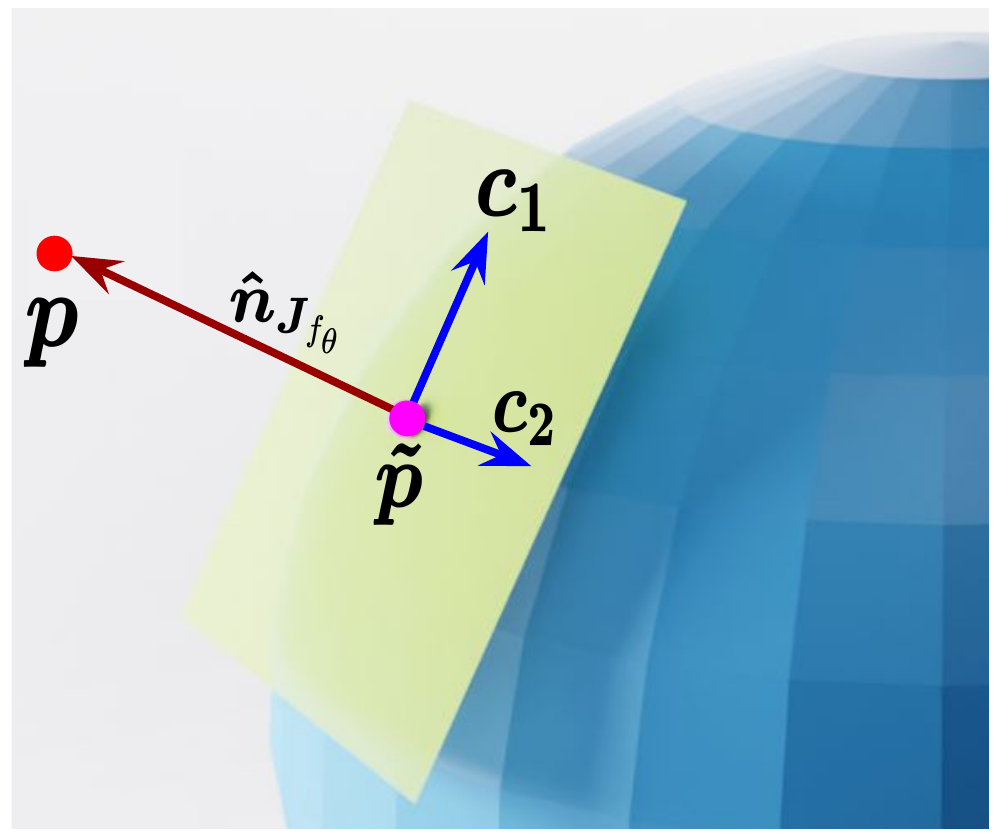}}\quad
  \caption{\textit{\textbf{Left:}} Network architecture of CSPNet (Sec.~\ref{sec:learning}). A point cloud $X$ is the input to the volume encoder $\psi$ to obtain a feature volume. The shape decoder $\phi$ is conditioned on it to obtain closest surface-point $\boldsymbol{\hat{p}}$ for each query point $\boldsymbol{p}\in\mathbb{R}^3$. Next, we show how \name enables extraction of both \textit{UDF}, \textit{NVF}, and tangent plane; which are further utilised for applications like rendering (via Sphere tracing in Sec.~\ref{sec:spheretracing}) and meshing (in Sec.~\ref{sec:meshing}). \textit{\textbf{Right:}} Surface-normal estimation via the method described in Sec.~\ref{sec:bwd_normals}. $\boldsymbol{c_1}$ and $\boldsymbol{c_2}$ refer to the basis of the tangent plane and $\boldsymbol{\hat{n}}_{{\boldsymbol{J}}_{f_\theta}}$ is the normal estimated.}
  \label{fig:sys_diag}
  \vspace{-4mm}

\end{figure*}

\noindent
\textbf{Explicit Shape Representations.} 
Explicit approaches primarily use voxels, meshes or point clouds for representing 3D shapes. Voxels provide a direct extension of pixels to 3D, allowing easier extension of image processing methods for 3D shape analysis. Several initial shape representation works are built upon this idea~\cite{voxnet-2015,choy2016,pix3d}. Drawbacks of using voxel representations are limited output resolution, and higher computational and memory requirements. Mesh representations address some of these issues \cite{atlasnet2018,wang2018pixel2mesh,meshrcnn-2019}, although the details captured are still limited by the (predetermined) choice for the number of vertices. Point clouds provide a more compact and sparser representation of surface geometry, but do not yield high-fidelity reconstruction and cannot reliably model arbitrary topology.

\noindent
\textbf{Implicit Shape Representations.} Modern implicit shape representation approaches use deep neural network models to implicitly represent a shape by either (1) classifying a (query) point as inside/outside a shape~\cite{occupancy_network}  (delineated by the modeled surface), or by (2) the signed (or unsigned) distance of the point to the surface~\cite{deepsdf}, where the sign indicates inside or outside.

Hybrid explicit/implicit representations are proposed \cite{chen2020bsp,cvxnet-2020}, where the implicit function is a union of inside/outside classifier hyper-planes. BSP-Net~\cite{chen2020bsp} uses a binary space partitioning network to model a convex decomposition of the 3D shape, the union of which defines a watertight separation of the inside/outside of the shape. CvxNet~\cite{cvxnet-2020}, also proposes a convex decomposition using 
hyper-planes but with a double representation of a complex primitive.

The methods described above ~\cite{chen2020bsp,cvxnet-2020,deepsdf,occupancy_network, deepsdf} can only represent closed surfaces, with an additional requirement that the training data also comprises of \textit{closed} watertight shapes, which often results in non-trivial loss of fine details~\cite{huang2018robust}. SAL \cite{sal-2020} provides a partial solution to this problem by proposing an unsigned similarity function to learn from ground truth unprocessed triangle soups, but eventually infers a shape representation which can only model closed shapes. NDF \cite{NEURIPS2020_f69e505b} overcomes this limitation by using a \textit{UDF} to represent both open and closed shapes, but cannot easily provide high fidelity estimates of differential geometry (surface normal, tangent plane) due to the vanishing gradient of the \textit{UDF} on the surface.

In contrast, \textit{\name}\textit{s} can model arbitrary topologies (both open and closed) while also allowing for simple, efficient and high-fidelity computation of differential surface geometry as opposed to the prior art~\cite{NEURIPS2020_f69e505b,sal-2020,chen2020bsp,occupancy_network,deepsdf,cvxnet-2020}. 

\section{Approach}
\label{sec:approach}


In this section we present the proposed shape representation, and how it can be used for downstream applications. A schematic of the system architecture is presented in Fig.~\ref{fig:sys_diag}. We start below with first defining the proposed implicit shape representation (Sec.~\ref{sec:shape_rep}) and deep neural network model for the same (Sec.~\ref{sec:learning}). Then, we present approaches to estimate the local geometric properties of the surface, e.g. the tangent plane and the surface normal (Sec.~\ref{sec:normals}) and  finally propose algorithms for using \name for downstream applications like rendering (Sec.~\ref{sec:spheretracing}) and meshing (Sec.~\ref{sec:meshing}).
\subsection{Shape Representation}
\label{sec:shape_rep}
Given a surface $\mathcal{S} \subset \mathbb{R}^3$ and a (query) point $\boldsymbol{p} \in \mathbb{R}^3$, we define the Closest Surface Point \name as a function $\name(\boldsymbol{p}): \mathbb{R}^3 \longmapsto \mathcal{S}$, such that 
\begin{equation}
    \begin{gathered}
        \name(\boldsymbol{p}) = \boldsymbol{\tilde{p}},\;s.t.\; \boldsymbol{\tilde{p}} = \operatorname*{argmin}_{\boldsymbol{p_s} \in \mathcal{S}} {\Vert\boldsymbol{p} - \boldsymbol{p_s}\Vert_2}
    \end{gathered}
\end{equation}
\noindent where $\boldsymbol{\tilde{p}}$ is the closest point on the surface $\mathcal{S}$ to the query point $\boldsymbol{p}$. Given the closest surface-point, \textit{UDF} can be trivially calculated as:
\begin{equation}
    \label{eq:udf}
    \textit{UDF}(\boldsymbol{p}) = \Vert\boldsymbol{p} - \boldsymbol{\tilde{p}}\Vert_2
\end{equation}

\subsubsection{CSPNet: A Deep Neural \name Model}
\label{sec:learning}
We model \name using a deep neural network which we demonstrate to be robust to training with noisy data generated from a noisy triangle soup. 
We illustrate the complete architecture in Fig.~\ref{fig:sys_diag}. There are two main components of the proposed CSPNet.\\




\noindent\textbf{Volume Encoder. } For any input 3D shape, the volume encoder $\psi$ produces a feature volume which is isotopic\footnote{Having the same topology as the enclosing volume (not the input 3D shape)} to the volume enclosing the input shape. Each feature voxel encodes properties of the surface from the vantage point of the voxel. For the implementation in this paper, we follow the architecture of Convolutional Occupancy Networks~\cite{10.1007/978-3-030-58580-8_31} for the volume encoder $\psi$. The encoder takes as input the entire point cloud for the input shape and produces a volumetric feature encoding. More specifically,  PointNet~\cite{qi2017pointnet} is used to encode point features. To get volumetric features, a voxel grid is constructed and voxel features are computed by (average) pooling features for the points that correspond to the voxel under consideration. This is followed by a 3D U-Net which produces the final encoding of the feature volume, resulting in a feature of dimension $F$ for each voxel. \\

\noindent\textbf{Shape Decoder. }The feature corresponding to the input query point $\boldsymbol{p}$ is sampled from the 3D feature volume using trilinear interpolation, and passed into the shape decoder $\phi$ along with query point $\boldsymbol{p}$. The shape decoder $\phi$ uses the features encoding the shape to predict the surface point closest to $\boldsymbol{p}$. Here on, we will use $f_{\theta}$ and $g_{\theta}$ to denote the DNN approximations to the \textit{\name} and the \textit{UDF} functions respectively, where $\theta$ denotes the union of parameters of both encoder and decoder. Note that the output of CSPNet, $f_{\theta}$, directly provides an estimate for \textit{\name} while the estimate for the \textit{UDF}, $g_{\theta}$, is obtained as $\|\boldsymbol{p}-f_\theta(\boldsymbol{p})\|_2$ using (\ref{eq:udf}).

\subsection{Differential Surface Geometry}
\label{sec:normals}

For any query point $\boldsymbol{p}$, CSPNet directly provides us with an estimate of both the closest point on the surface $f_\theta(\boldsymbol{p})$ as well as the unsigned distance to it, $g_\theta(\boldsymbol{p})$. However, in addition, a variety of downstream applications in vision~\cite{kazhdan2006poisson, pomerleau2015review}, robotics~\cite{robot_duhautbout2019distributed, robot_oleynikova2016signed}, graphics~\cite{correa2003out}, and animation~\cite{volino1999fast} need estimates of local differential properties of the surface like the tangent plane and normal at any surface point. We show below how we can easily estimate these properties.    


\subsubsection{Using the Jacobian}
\label{sec:bwd_normals}
Let $\boldsymbol{p}$ be any query point and $\boldsymbol{\tilde{p}} \in \mathcal{S}$ be its closest point on the surface $\mathcal{S}$. Further, let $\boldsymbol{J}_{f_{\theta}}(\boldsymbol{\tilde{p}})$ denote the Jacobian of $f_{\theta}$ at $\boldsymbol{\tilde{p}}$. Let $\delta$ be the unsigned distance from $\boldsymbol{p}$ to $\boldsymbol{\tilde{p}}$ and $\boldsymbol{d}$ be the surface normal at $\boldsymbol{\tilde{p}}$. Then, we get the following approximation using the first-order Taylor series expansion:  
\begin{equation}
    \begin{gathered}
        \boldsymbol{p} = \boldsymbol{\tilde{p}} + \delta \cdot \boldsymbol{d}\\
        f_{\theta}(\boldsymbol{p}) = f_{\theta}(\boldsymbol{\tilde{p}} + \delta \cdot \boldsymbol{d})\\
        \boldsymbol{\tilde{p}} \approx \boldsymbol{\tilde{p}} + \delta \cdot \boldsymbol{J}_{f_{\theta}}(\boldsymbol{\tilde{p}}) \cdot \boldsymbol{d}\\
        0 \approx \delta \cdot \boldsymbol{J}_{f_{\theta}}(\boldsymbol{\tilde{p}}) \cdot \boldsymbol{d}
    \end{gathered}
\end{equation}

The last equation shows that (to a first order approximation of the surface), the surface normal $\boldsymbol{d}$ lies in the null space of the Jacobian $\boldsymbol{J}_{f_{\theta}}(\boldsymbol{\tilde{p}})$ while the span of the Jacobian provides the tangent space of the surface. This is illustrated in the Fig.~\ref{fig:sys_diag}{\color{red}b} and is intuitively clear since along the  direction perpendicular to the surface, the CSP function does not change, giving the same closest surface point. The tangent space and the normal to the surface both can be estimated using singular value decomposition (SVD). 

However, computation of Jacobian requires a backward pass through CSPNet. Prior works which differentiate the distance function on the zero level-set (i.e the surface)~\cite{deepsdf} also need a backward pass. Even so, since the derivative of \textit{UDFs} vanish at the surface, NDF estimates the normals close to the surface~\cite{NEURIPS2020_f69e505b} leading to some loss in fidelity.



\subsubsection {Forward Mode Normal Estimation}
\label{sec:fwd_normals}
In certain applications like rendering, sphere tracing is used to obtain a point on the surface and it is needed to quickly and efficiently estimate the normal at the point of intersection~\cite{correa2003out}. We can use the Jacobian approach presented in the previous section but it requires a backward pass.  

An alternate approach for obtaining a fast approximation for the surface normal, using a forward pass from a query point $\boldsymbol{p}$ close to but not on the surface is by using the Normal Vector Field (\textit{NVF}) defined as follows: 
\vspace{-2mm}
\begin{equation}
    \label{eq:nvf}
    \textit{NVF}(\boldsymbol{p}) = \frac{\boldsymbol{p} - \boldsymbol{\tilde{p}}}{\textit{UDF}(\boldsymbol{p})}
\end{equation}

\noindent We represent the corresponding estimate for \textit{NVF} by $h_\theta$ as $(\boldsymbol{p} - f_\theta(\boldsymbol{p}))/g_\theta(\boldsymbol{p})$. We refer to this method of estimating normals as forward-mode normal estimation. Since there is no backward pass involved, it is faster than the previous methods. We demonstrate the utility of this approach in Sec.~\ref{sec:local surface prop} and validate its performance both in terms of accuracy and speed via extensive experimental evaluation. More generally, fast estimation of the \textit{NVF} at off-surface locations is vital to robotics applications such as path planning in distance fields~\cite{robot_duhautbout2019distributed, robot_oleynikova2016signed}.

\subsection{Rendering and Meshing}
\label{sec:MISE}
In this section, we describe techniques for rendering surfaces and extracting topologically consistent meshes from the the learnt representation. Note that this process is important for many downstream vision applications such as shape analysis~\cite{laga20183d} and graphics applications such as rendering novel scenes under changed illumination, texture or camera viewpoints~\cite{purcell2005ray}. 

\vspace{-2mm}
\subsubsection{Sphere Tracing \textit{\name}} 

\begin{figure}
    \centering
    \includegraphics[width=\columnwidth]{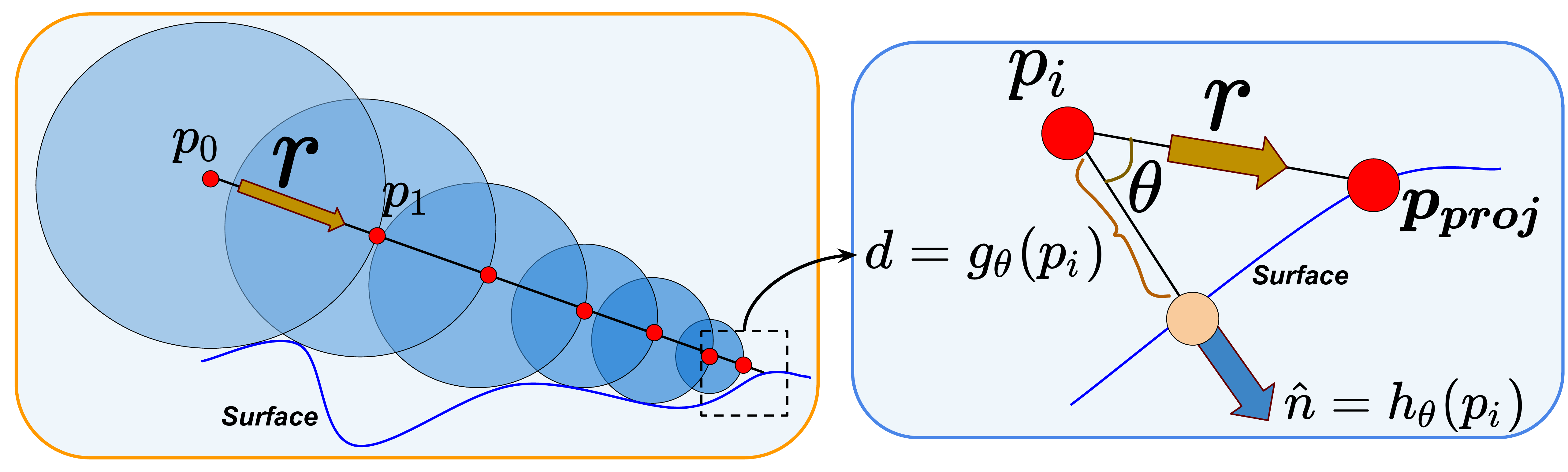}
    \caption{\textit{\textbf{Left:}} An illustration of the Sphere Tracing procedure described in Section.~\ref{sec:spheretracing}. \textit{\textbf{Right:}} Leveraging the \textit{NVF} for obtaining more accurate ray-scene intersections.}
    \label{fig:sphere_trace}
    \vspace{-5mm}
\end{figure}


\label{sec:spheretracing}Sphere tracing~\cite{hart1996sphere} is a standard technique to render images from a distance field that represents the shape. To create an image, rays are cast from the focal point of the camera, and their intersection with the scene is computed using sphere tracing. Roughly speaking, irradiance/radiance computations are performed at the point of intersection to obtain the color of the pixel for that ray. 

The sphere tracing process can be described as follows: given a ray, $\boldsymbol{r}$, originating at point, $\boldsymbol{p_0}$, iterative marching along the ray is performed to obtain its intersection with the surface. In the first iteration, this translates to taking a step along the ray with a step size of $\textit{UDF}(\boldsymbol{p_0})$ to obtain the next point $\boldsymbol{p_1} = \boldsymbol{p_0} + \boldsymbol{r} \cdot g_{\theta}(\boldsymbol{p_0})$. Since $g_{\theta}(\boldsymbol{p_0})$ is the smallest distance to the surface, the line segment $[\boldsymbol{p_0}, \boldsymbol{p_1}]$ of the ray is guaranteed not to intersect the surface ($\boldsymbol{p_1}$ can touch but not transcend the surface). The above step is iterated $i$ times till $\boldsymbol{p_i}$ is $\epsilon$ close to the surface. The i-th iteration is given by $\boldsymbol{p_i} = \boldsymbol{p_{i-1}} + g_{\theta}(\boldsymbol{p_{i-1}})$ and the stopping criteria $g_{\theta}(\boldsymbol{p_i})\leq\epsilon$. 

Note that the above procedure can be used to get close to the surface but does not obtain a point on the surface. Once we are close enough to the surface, we can use a local planarity assumption (without loss of generality) to obtain the intersection estimate. This is illustrated in Figure~\ref{fig:sphere_trace} and is obtained in the following manner:  if we stop the sphere tracing of the \textit{\name} at a point $\boldsymbol{p_i}$, we evaluate the \textit{NVF} at that point as $\boldsymbol{\hat{n}} = h_{\theta}(\boldsymbol{p_i})$, and compute the cosine of the angle between the \textit{NVF} and the ray direction. The estimate is then obtained as $\boldsymbol{p_{proj}} = \boldsymbol{p_i} + \boldsymbol{r} \cdot \frac{g_{\theta}(\boldsymbol{p_i})}{\boldsymbol{r}^{T}\boldsymbol{\hat{n}}}$. 



\vspace{-5mm}
\subsubsection{From \textit{\name} to Meshes}
\label{sec:meshing}
Sphere tracing \textit{\name}, described in the previous section, can only be used to render a view of the shape. Thus, the extracted surface is immutable and cannot be used for applications such as 3D shape modeling, analysis and modification ~\cite{laga20183d}. Explicit 3D mesh representations are more amenable for such applications. In this section, we propose an approach to extract a 3D mesh out of the learnt \textit{\name}.

A straightforward way to extract a mesh from an implicit representation is to create a high-resolution 3D distance grid and using the marching cubes algorithm~\cite{marchingcubes} on this grid. However, as discussed in~\cite{occupancy_network} this process is computationally expensive at high-resolutions, as we need to densely evaluate the grid. In~\cite{occupancy_network} a method for multi-resolution surface extraction technique is proposed by hierarchically creating a binary occupancy grid by conducting inside/outside tests for a binary classifier based implicit representation. 

\begin{figure}
    \centering
    \includegraphics[scale=0.125]{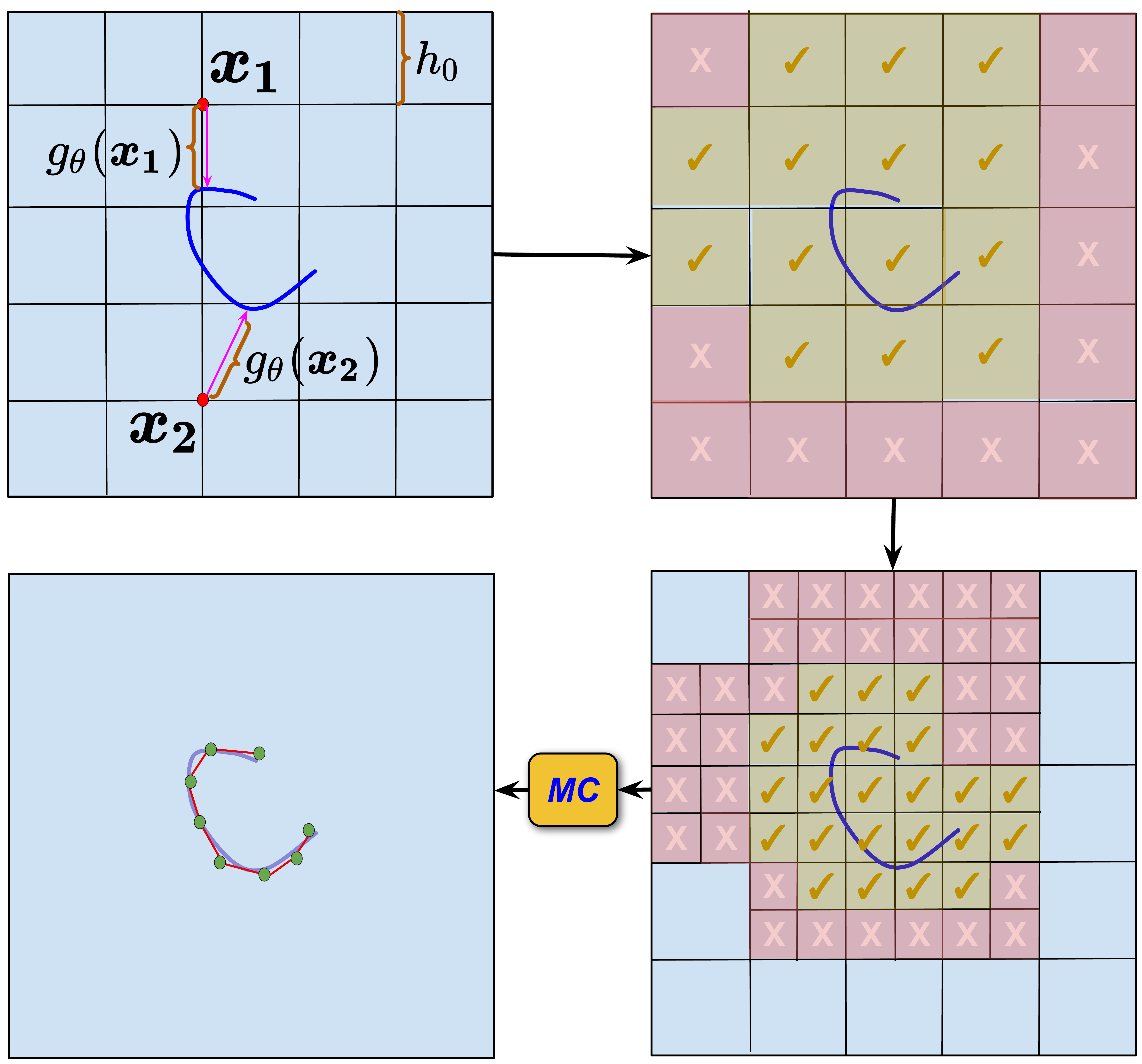}
    \caption{Multi-Resolution surface extraction for \textit{\name} (described in Section.~\ref{sec:meshing}). At each level of hierarchy, we show the voxels selected for subdivision. Note that there are a some false positive voxels selected close to the shape which get eliminated in the next hierarchical level (Step 2 to Step 3), and are pruned out by using a small positive threshold while meshing the distance-grid in the Marching cubes (MC) step. }
    \label{fig:marching_cubes}
    \vspace{-4mm}
\end{figure}

However, \textit{\name}\textit{s} cannot perform inside/outside tests. Hence we propose a novel technique to hierarchically divide the distance grid using edge lengths of the voxel grid cubes. We illustrate the procedure in Fig.~\ref{fig:marching_cubes}. Starting with a voxel grid at some initial resolution, we obtain a high resolution distance grid and perform marching cubes on the grid using a small positive threshold to get the final mesh. 
A voxel is chosen for subdivision if any of its eight corners have the predicted \textit{UDF} value $g_{\theta}(\boldsymbol{x}) < h_i$, where $h_i$ is the edge length of the voxel grid at the $i'th$ level. The voxels that are not chosen for subdivision are simply discarded in the next level. Using this procedure, we quickly obtain a high-resolution distance grid, which is converted to a mesh using marching cubes. Note that, our algorithm selects a few false positive voxels in the final resolution, but these are effectively pruned out in the final mesh by using a small positive threshold in the marching cubes~\cite{marchingcubes} step.     



\section{Experiments}
\label{sec:experiments}
In this section, we validate the different parts of our proposed system outlined in Fig.~\ref{fig:sys_diag} against a selection of prior art. First we demonstrate the superiority of the proposed implicit shape representation (\textit{\name}: Sec.~\ref{sec:shape_rep}) on the task of surface reconstruction from point clouds. Next, we validate the proposed methods for extracting local surface properties such as surface normals (described in Sec.~\ref{sec:normals}). Finally, we test the novel sphere-tracing algorithm for \textit{\name}\textit{s} and the coarse to fine meshing algorithm (described in Sec.~\ref{sec:MISE}). 

\vspace{1mm}

\noindent
\textbf{Baselines.} Most existing methods such as Occupancy Networks~\cite{occupancy_network}, DeepSDF~\cite{deepsdf}, Point-Set Generation Networks~\cite{psgn}, Deep Marching Cubes~\cite{Li:2018:DMC} and IF-Net~\cite{chibane2020implicit} only work on watertight (i.e. closed) shapes. We compare against these methods to verify that our method retains performance on closed shapes, in addition to being able to model open shapes. On the other hand, for comparing performance on raw/unprocessed shapes, we choose SAL~\cite{sal-2020} and NDF~\cite{NEURIPS2020_f69e505b}. While these methods can work with non-watertight (i.e. unprocessed) ground truth, they still require a backward pass to estimate surface normals, which leads to an added computational and memory footprint. Additionally, while NDF can reconstruct both open as well as closed shapes, it is unable to guarantee plane reproduction and accurate normal estimates on the surface of the shape. In the following sections, we show empirically that our method addresses these challenges.



\subsection{Shape Representation}
\label{sec:shape space}
In this section, we demonstrate the representational power of our model on ShapeNet dataset~\cite{chang2015shapenet}. We consider the task of surface reconstruction from point clouds, and first evaluate on closed shapes to verify that our proposed generic shape representation yields comparable performance to state-of-the art methods which are solely meant to work on watertight shapes~\cite{deepsdf, occupancy_network, psgn}. Second, we evaluate our method against NDF and SAL on raw, unprocessed shapes. Before describing the results, we present the evaluation metrics that we consider.

\vspace{-1mm}
\subsubsection{Evaluation metrics}
\label{sec:eval metrics}

A common practice for evaluating 3D reconstruction pipelines is the chamfer distance metric~\cite{deepsdf, sal-2020, NEURIPS2020_f69e505b}. However, as discussed in some prior work~\cite{chamfer_style}, this metric does not reflect the perceptual quality of the rendered image. Moreover, for applications such as relighting~\cite{maier2017intrinsic3d} it is desirable to obtain surface normal maps by directly rendering the iso-surface using sphere-tracing, as opposed to extracting a mesh. Clearly, there is a need to evaluate implicit shape representations on the perceptual quality of their iso-surfaces rendered via sphere-tracing. Therefore, in addition to the chamfer distance we propose new metrics (outlined below) which are designed to reliably capture these properties.   \\ 

\noindent
\textbf{Depth Error (DE).} First we evaluate the mean absolute error (\textit{MAE}) between the ground truth and the estimated depth map obtained by sphere-tracing the learnt representation. This error is evaluated only on the ``valid" pixels, which we define as the pixels having non-infinite depth (foreground) in both the ground truth and estimated depth map. This metric captures the \textit{accuracy of ray-surface intersection}.\\

\noindent
\textbf{Normal Cosine Similarity (NCS).} We also evaluate the cosine similarity between the sphere-traced normal map and the ground truth normal map for the valid pixels. Since the surface normals play a vital role in rendering, this metric is informative of the \textit{fidelity} of the rendered surface. \\
    
\noindent
\textbf{Pixel-Space IOU.} Finally, since both Depth Error and NCS are evaluated only on the valid pixels, they do not quantify whether the \textit{geometry of the final shape} is correct. Therefore, we also evaluate Pixel-Space IOU,
    \begin{equation}
        IOU = \frac{\#\textit{Valid}\, \textit{Pixels}}{\#\textit{Invalid}\, \textit{Pixels} + \#\textit{Valid}\, \textit{Pixels}}
    \end{equation}
Here the invalid pixels are those which have non-infinite depth (foreground) in either the ground truth depth map or the estimated depth map but not both. Note that for the proposed metrics, we render the shape from 6 views (uniformly sampled on sphere) to capture all the regions of the surface.

\vspace{-4mm}

\subsubsection{Data creation}
\label{sec:data creat}
We normalize each mesh in the ShapeNet dataset to $[-0.5, 0.5]$. For each shape, we densely sample a set of 0.25M points, denoted by the set $\mathcal{V}$, to represent the set of surface points. The training points $\mathcal{P}$ are obtained for each shape by uniformly sampling 0.025M points as well as perturbing the set $\mathcal{V}$ with a gaussian noise of \num{2.5e-4} and \num{2.5e-3}. Finally, the ground truth for each of these training points $\boldsymbol{p} \in \mathcal{P}$ is computed by finding the nearest surface point  $\boldsymbol{\tilde{p}} \in \mathcal{V}$ to construct the training pair $(\boldsymbol{p}, \boldsymbol{\tilde{p}})$. 

\subsubsection{Training}

Note that, we only train $f_{\theta}$, and $g_{\theta}$ and $h_{\theta}$ can be derived from it in the same way \textit{UDF} and \textit{NVF} can be derived from \textit{\name} in eq.~\ref{eq:udf} and eq.~\ref{eq:nvf} respectively. Given $f_{\theta}(\boldsymbol{p}|X) = \phi(\psi(X), \boldsymbol{p})$,
as the training objective, we simply use the squared $L_2$ loss between the estimated closest surface-point $f_{\theta}(\boldsymbol{p}|X)$ and the ground-truth $\boldsymbol{\tilde{p}} (=\name(\boldsymbol{p}|X))$
\begin{equation}
    \mathcal{L}_{\name} = \frac{1}{{\vert\mathcal{P}\vert}}\sum_{\boldsymbol{p}\in\mathcal{P}}||f_{\theta}({\boldsymbol{p}|X}) - \boldsymbol{\tilde{p}}||^{2}_{2} 
    \label{eq:l2}
\end{equation}


\noindent
\subsubsection{Evaluation on closed shapes}
We convert all the ShapeNet 3D models to closed shapes by following the steps in~\cite{stutz2018learning}. Following this, we run our data creation process (outlined in Sec.~\ref{sec:data creat}). After training our proposed surface reconstruction pipeline, we compare to the selected prior art outlined earlier and report the results in Table.~\ref{tab:exp_closed_shape}. Note that DeepSDF trains separate models for different classes in ShapeNet, whereas NDF, Occupancy Networks and PSGN train class-agnostic models. Hence, we train a class-agnostic \textit{\name} and finetune the model on individual classes to report results in Table~\ref{tab:exp_closed_shape}. We find that our per-class models outperform DeepSDF, and our class agnostic model performs on par with NDF.

 \begin{table}[h!]
 \resizebox{\columnwidth}{!}{
 \begin{tabular}{ cc } 
 
 \begin{tabular}{cc}
    \toprule
    Method & Chamfer-\textit{L2} $\downarrow$ \\
    \midrule
    PSGN~\cite{psgn} & 4.0\\
    Occ. Net~\cite{occupancy_network} & 4.0 \\
    DMC~\cite{Li:2018:DMC} & 1.0 \\
    IF-Net~\cite{chibane2020implicit} & 0.2 \\
    NDF~\cite{NEURIPS2020_f69e505b} & \textbf{0.05} \\
    \name (Ours) & 0.1 \\
     \bottomrule
 \end{tabular} &

 \begin{tabular}{ccc}
    \toprule
    \multirow{2}{*}{Class} & \multicolumn{2}{c}{Chamfer-\textit{L2} $\downarrow$} \\
    \cmidrule{2-3} 
    & DeepSDF~\cite{deepsdf} & \name (Ours) \\
    \midrule
    Sofas & 3.29 & \textbf{0.32}  \\
    Chairs & 3.41 & \textbf{0.15}  \\
    Tables & 8.39 & \textbf{0.20}  \\
    Planes & 1.77 & \textbf{0.04}  \\
    Lamps & 9.09 & \textbf{0.09}  \\
     \bottomrule
 \end{tabular} \\
 \end{tabular}
 }
 \caption{Results on closed shapes. \textbf{\textit{Left:}} Single model for all classes of ShapeNet. \textit{\textbf{Right:}} Comparison to DeepSDF~\cite{deepsdf} on per-class models. Note that the numbers are reported ($\times10^{4}$).}
  \label{tab:exp_closed_shape}
 \end{table}

\vspace{-3.5mm}
\subsubsection{Evaluation on unprocessed shapes}
In addition to closed shapes, \textit{\name} can also represent shapes of arbitrary topology. Therefore, we also train on unprocessed ShapeNet 3D models and evaluate performance using the metrics defined in Sec.~\ref{sec:eval metrics}. We compare against SAL and NDF, which are methods that can learn representations from raw/unprocessed ground truth. \footnote[1]{Since both SAL and NDF do not provision a release of pretrained class-agnostic models, we retrain them using code provided by authors.} This comparison is reported in Table~\ref{tab:exp_open_shape}. We find that \textit{\name} marginally outperforms NDF on chamfer, depth and IOU metrics, but yields a significant improvement on the surface normals metric owing to the useful plane reproduction property of \textit{\name}. Additionally, SAL clearly suffers on all metrics, given that it learns a signed distance function (closed shape) even for surfaces that are open. This behavior can also be confirmed in the qualitative results shown in row 2 of Fig.~\ref{fig:shape_recon}.

\begin{table}[h!]
 \resizebox{\columnwidth}{!}{

 \begin{tabular}{ccccc}
    \toprule
    Method & Chamfer-\textit{L2} $\downarrow$ & Depth $\downarrow$ & Normal $\uparrow$ & IOU $\uparrow$ \\
    \midrule
    SAL~\cite{sal-2020} & 2.25 & 0.025 & 0.84 & 0.96\\
    NDF~\cite{NEURIPS2020_f69e505b} & 1.73 & 0.018 & 0.86 & 0.97 \\
    \textit{\name (Ours) }& \textbf{1.28} & \textbf{0.014} & \textbf{0.92} & \textbf{0.98}\\
     \bottomrule
 \end{tabular} 
 
 }
 \caption{Results on unprocessed shapes. We evaluate both on chamfer distance metric as well as the three additional metrics defined in Sec.~\ref{sec:eval metrics}. For all methods, we obtained normals by leveraging first order information. Chamfer distance is reported ($\times10^4$).}  \label{tab:exp_open_shape}
 \end{table}

\vspace{-4mm}

\subsection{Local Surface Properties}
\label{sec:local surface prop}
In Sec.~\ref{sec:normals}, we described various strategies to estimate surface normals using the learnt implicit representation. 
We refer to the strategy using the Jacobian as \textit{\name (jac.)} and the one using the forward pass 
(eqn.~\ref{eq:nvf}) 
as \textit{\name (fwd.)}. 

Similar to NDF, this latter approach approximates surface normals using off-surface points close to the surface (where $\boldsymbol{p} \approx \boldsymbol{\tilde{p}}$) by stepping back along the ray at its point of intersection with the surface. More concretely, given a ray $\boldsymbol{r}$ which intersects with the surface at $\boldsymbol{p_{int}}$ (at the end of sphere-tracing), the normal is computed by stepping back along the ray by some scalar value $\alpha$. Thus, $\hat{n}_{\boldsymbol{p_{int}}} = \nabla_{\boldsymbol{p_{int}}}{g_\theta(\boldsymbol{p_{int}} - \alpha \cdot \boldsymbol{r})}$. Note here that $\alpha$ is a hyperparameter which is sensitive to the curvature of the surface, and NDF chooses a constant $ \alpha = 0.005$. However, we observe that choosing a single $\alpha$ for all shapes is sub-optimal given that surfaces can have varying curvatures. 

To investigate the sensitivity of the system to varying $\alpha$, we plot normal cosine similarity vs different values of $\alpha$ in Table~\ref{tab:normal_est}. It can be clearly seen that \textit{\name (jac.)} has higher quality normal estimates for points on the surface (i.e. $\alpha = 0$), given its tangent plane reproduction property, as opposed to NDF and \textit{\name (fwd.)} which do not. It is interesting to note that although SAL learns a signed distance function that is differentiable on the surface of the shape, it still performs poorly on this metric, owing to the instability of their unsigned similarity loss, and poor geometric reconstruction on open shapes. However, we find that both NDF and \textit{\name (fwd.)} yield comparable performance to \textit{\name (jac.)} if allowed to step back along the ray ($\alpha = 0.005$). However, the normal cosine similarity is lower than \textit{\name (jac.)} at $\alpha = 0$, which is a definite drawback. 
Moreover, we find that $\alpha = 0.005$, \textit{\name (fwd.)} yields similar performance compared to NDF, even though it does not use a backward pass. We report rendering speeds and memory footprint for \textit{\name (fwd.)} and NDF in Table~\ref{tab:speed_memory_rendering}, and we immediately find that \textit{\name} is superior on both fronts.

\begin{figure}
\vspace{-1mm}
    \centering
    \includegraphics[width=\columnwidth]{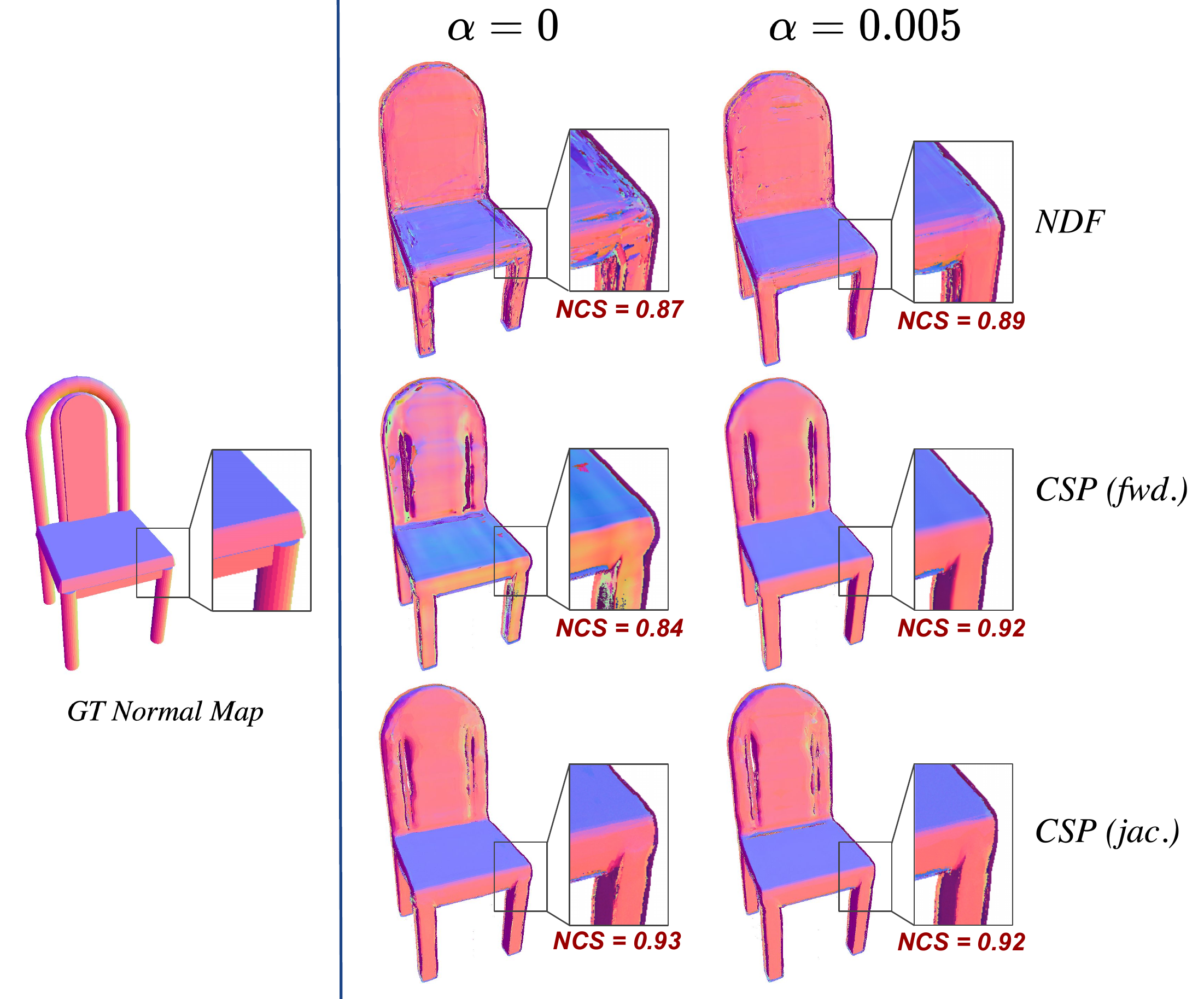}
    \caption{Comparison of normal estimation strategies outlined in Sec.~\ref{sec:normals}. Note that \textit{\name (jac.)} at $\alpha=0$  is the best performing model (0.93), with \textit{\name (fwd.)} at $\alpha=0.005$ following next (0.92).}
    \label{fig:normals}
    \vspace{-4mm}
\end{figure}

Additionally, in Table~\ref{tab:normal_est} we find that although $\alpha = 0.005$ yields reasonably good normals, the standard deviation is higher than those obtained by the tangent plane approximation. This clearly shows that choosing a single threshold for all shapes~\cite{NEURIPS2020_f69e505b} is sub-optimal. Finally, we qualitatively compare various normal estimation strategies in Fig.~\ref{fig:normals}. We find here too that \textit{\name (fwd.)} performs reasonably well for $\alpha = 0.005$, with \textit{\name (jac.)} yielding the best performance at $\alpha = 0$. Both visually and quantitatively, we find that our normal estimation strategies outperform NDF. Additionally, forward-mode surface estimates, \textit{Ours (fwd.)} are faster than that of NDF while \textit{Ours (jac.)} is comparable in speed (more analysis in supplementary).


\begin{table}[h!]
 \resizebox{\columnwidth}{!}{
 \begin{tabular}{ccccc}
    \toprule
    \multirow{2}{*}{Method} & \multicolumn{4}{c}{Normal Map Similarity} \\
    \cmidrule{2-5}
    & $\alpha=0$ & $\alpha=0.002$ & $\alpha=0.005$ & $\alpha=0.05$ \\ 
    \midrule
    SAL~\cite{sal-2020} & 0.84 $\pm$ 0.017 & 0.851 $\pm$ 0.014 & 0.871 $\pm$ 0.009 & 0.861 $\pm$ 0.01 \\
    NDF~\cite{NEURIPS2020_f69e505b} & 0.863 $\pm$ 0.01 & 0.882 $\pm$ 0.008 & 0.903 $\pm$ 0.006 & 0.891 $\pm$ 0.008 \\
    \hline
    \textit{\name (fwd.)} & 0.620 $\pm$ 0.12 & 0.873 $\pm$ 0.018 & 0.912 $\pm$ 0.006 & \textbf{0.91 $\pm$ 0.007} \\
    \textit{\name (jac.)} & \textbf{0.913 $\pm$ 0.003}  & \textbf{0.915 $\pm$ 0.003} & \textbf{0.920 $\pm$ 0.003} & 0.871 $\pm$ 0.01 \\
     \bottomrule
 \end{tabular} 
 }
 \vspace{-0.05mm}
 \caption{Normal estimation accuracy of various methods described in Sec.~\ref{sec:normals}. 
 Here $\alpha$ refers to the step-back distance along the ray.}  \label{tab:normal_est}
 \end{table}

\vspace{-4mm}
\subsection{Rendering and Meshing}
\label{sec:rend_mesh}
In this section, we validate our sphere-tracing strategy and meshing algorithm (Sec.~\ref{sec:MISE}) against various baselines.

\noindent

\begin{figure}
    \centering
    \includegraphics[width=\columnwidth]{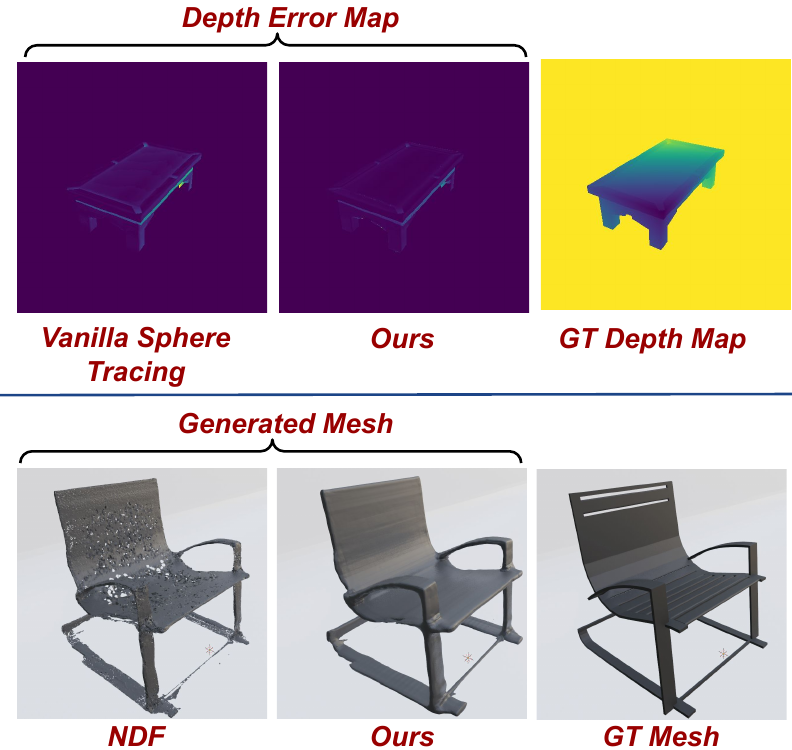}
    \caption{\textit{\textbf{Top:}} Depth error maps comparing vanilla sphere tracing strategy vs our projection based strategy 
    \textit{\textbf{Bottom:}} Comparison of our meshing algorithm with that of NDF. Note that NDF displays visible artifacts, whereas our strategy reconstructs a topologically consistent mesh.}
    \label{fig:depth_error}
    \vspace{-4mm}
\end{figure}

\textbf{Sphere Tracing \textit{\name}.} We compare the sphere tracing strategy described in Sec.~\ref{sec:spheretracing} to a baseline strategy when the projection step is excluded from the algorithm. Our proposed strategy yields better depth maps (\textit{MAE} = $0.014$) than the Vanilla Sphere tracing (\textit{MAE} = $0.016$) owing to more accurate ray-scene intersection. As expected, the qualitative results (depth error maps) shown in Fig.~\ref{fig:depth_error}  also indicate the benefit of using projection step as a part of sphere tracing \textit{\name}. Refer supplementary material for more visualizations.

\noindent
\textbf{Speed \& memory footprint of rendering.} In Table~\ref{tab:speed_memory_rendering}, we report the average time taken to render a 512$\times$512 image using a memory budget of 8GiB. Since we do not rely on backward passes through the network (see definition of $\textit{NVF}$ in Sec.~\ref{sec:fwd_normals}) higher batch sizes can be used on a fixed GPU budget, which leads to 20$\times$ faster rendering. In this manner, \textit{\name} provides a viable solution for applications which require real time, fast estimation of surface normals on small GPUs with limited memory~\cite{robot_duhautbout2019distributed, robot_duhautbout2019distributed, correa2003out}. Refer to supplementary material for more details on the specifics of the experimental setup.

\begin{table}[h!]
\centering
 \resizebox{0.82\columnwidth}{!}{
\begin{tabular}{cc}



 \begin{tabular}{cccc}
    \toprule
    Method & \pbox{10cm}{ \relax\ifvmode\centering\fi Rendering \\ time} & \pbox{10cm}{ \relax\ifvmode\centering\fi \#decoder \\ params} & \pbox{10cm}{\relax\ifvmode\centering\fi Memory \\ Budget}  \\
    \midrule
    NDF  & 0.063s & 1.97M & \multirow{2}{*}{8GiB} \\
    \textit{\name (fwd.)} & \textbf{0.003s} & 1.91M & \\
     \bottomrule
 \end{tabular} \\
 \end{tabular} 
 }
 \vspace{0.6mm}
 \caption{Rendering times for a 512$\times$512 image and memory footprint of the proposed method (\textit{\name}) against NDF.} 
 \label{tab:speed_memory_rendering}
 \vspace{-4mm}
 \end{table}


\noindent
\textbf{Meshing \textit{\name}.} Our novel course-to-fine meshing algorithm allows for fast conversion of \textit{\name} to mesh, providing a viable and fast alternative to that proposed in NDF~\cite{NEURIPS2020_f69e505b}. For the representative example shown in Fig.~\ref{fig:depth_error} NDF's method takes 4.48s to generate the dense point cloud and an additional 104s for meshing using BPA~\cite{bernardini1999ball}\footnote{BPA is known to be slow~\cite{bernardini1999ball}}. On the other hand, our method takes a total of 2.50s for generating the final mesh. It is also important to note that although we use only 636k function evaluations, a higher quality mesh is recovered in comparison to NDF which uses 12M function evaluations. In Fig.~\ref{fig:depth_error} we show qualitative comparisons to NDF's meshing algorithm for \textit{UDFs}. Note also that NDF's meshes display visible artifacts, given that it recovers the mesh after performing BPA on a dense point cloud (1M pts.) generated from the learnt  representation. In contrast, our coarse-to-fine meshing strategy enables the application of Marching Cubes, and reconstructs a topologically consistent and visually pleasing mesh. Refer supplementary for details on the hyperparameters used. 

 
\section{Conclusion}
In this work, we proposed a new class of implicit representations called \name that can model complex 3D objects (both open and closed surfaces), with a fidelity surpassing the state of the art. We demonstrated that \name also facilitates accurate and efficient computation of local geometric properties of the surface like the tangent plane and the surface normal which enables efficient algorithms for downstream applications like surface rendering and meshing - we presented novel algorithms for both. We further showed that \name yields state-of-the-art performance on the unprocessed ShapeNet dataset, surpassing prior art such as SAL~\cite{sal-2020} and NDF~\cite{NEURIPS2020_f69e505b}. In summary, this work provides a strong alternative to existing methods for 3D modeling and representation by addressing fundamental problems in representing complex shapes. In the future, we expect to extend this work to infer surface representations - both geometric and photometric - from single and multi-view 2D images.





\section{Acknowledgements}
This research was supported in part by NSF award No. IIS-1925281.

{\small
\bibliographystyle{ieee_fullname}
\bibliography{egbib}
}

\clearpage

\appendix
\onecolumn
\section*{Supplementary Material}

In the paper, we presented results from the shape-agnostic CSP network (a single function for all shapes) which for a given encoded shape provided at the input, produced the closest surface point for the queried input point. Here, we first supplement those results with a single-shape CSP network. This is presented in (Sec.~\ref{sec:single_shape}) below. 

Subsequent sections present additional details for the experimental evaluation in the main paper as follows: 
\begin{itemize} [noitemsep]
\item The network architecture and training details pertaining to the shape representation trained in Section~\ref{sec:shape space} in the main paper are presented in Sec.~\ref{sec:train_arch}.
\item Details for the Jacobian computed in Section~\ref{sec:local surface prop} of the main paper and its computational performance are presented in Sec.~\ref{sec:speed}.
\item The experimental setup used for rendering and meshing (Section~\ref{sec:rend_mesh} of the main paper) is described next in (Sec.~\ref{sec:rend_mesh_setup}). 
\item Sec.~\ref{sec:add_qual_res} presents additional qualitative results for rendering via sphere tracing, supplementing those presented in Section~\ref{sec:rend_mesh} of the main paper. 
\item Sec.~\ref{sec:tools} share the details of the various off-the-shelf tools used in our implementations and experimental evaluation.
\end{itemize}


\section{Single-shape \name}
\label{sec:single_shape}

\begin{wrapfigure}{r}{0.15\textwidth}
  \begin{center}
    \includegraphics[scale=0.1]{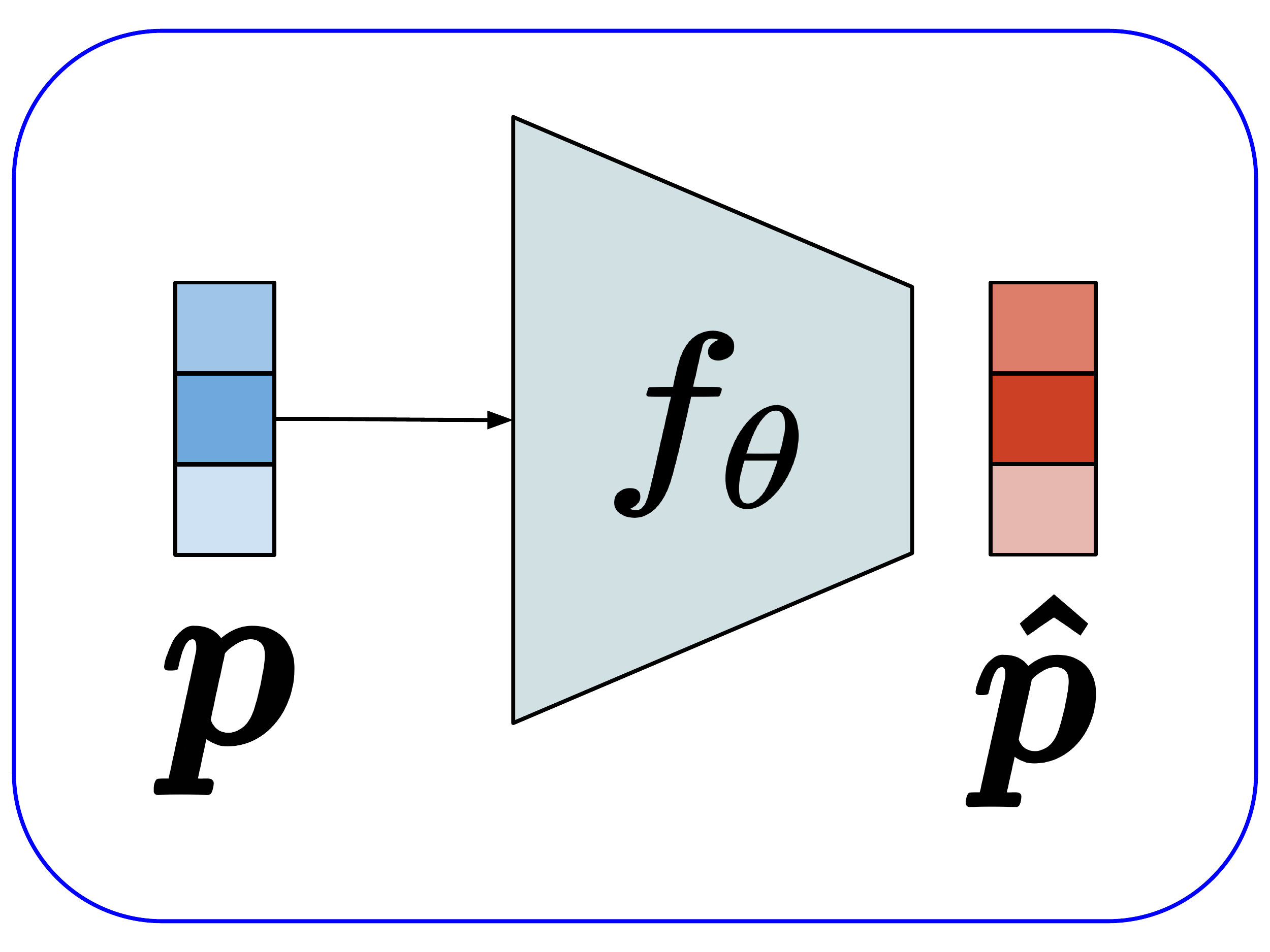}
  \end{center}
  \vspace{-5mm}
  \caption{Single shape \name}
  \label{fig:arch_sing}
\end{wrapfigure}

While the primary focus of this work was to build a single, shape-agnostic \name model, we present here a model for a single-shape \name implemented as follows: For any input point (or query point) in the 3D space, $\boldsymbol{p}$, a 10-layer MLP estimates the closest point on the surface, $\boldsymbol{\hat{p}}$. Let \codeword{fci} denote a fully-connected layer with \codeword{i} output dimensions. Then the MLP is given by \\
\vspace{-5mm}
\begin{center}
    {\codeword{fc120, fc512, fc1024, fc2048, fc2048,} \\
    \codeword{fc1024, fc512, fc256, fc128, fc3}}
\end{center}
\noindent where the input dimension of \codeword{fci} is determined by the output dimension of the layer prior to it and every \codeword{fc} layer is followed by $ReLU$ non-linearity, except the final layer. The  architecture of the single-shape \name is presented in Fig.~\ref{fig:arch_sing}.

\par
\begin{figure*}
    \centering
    \includegraphics[scale=0.20]{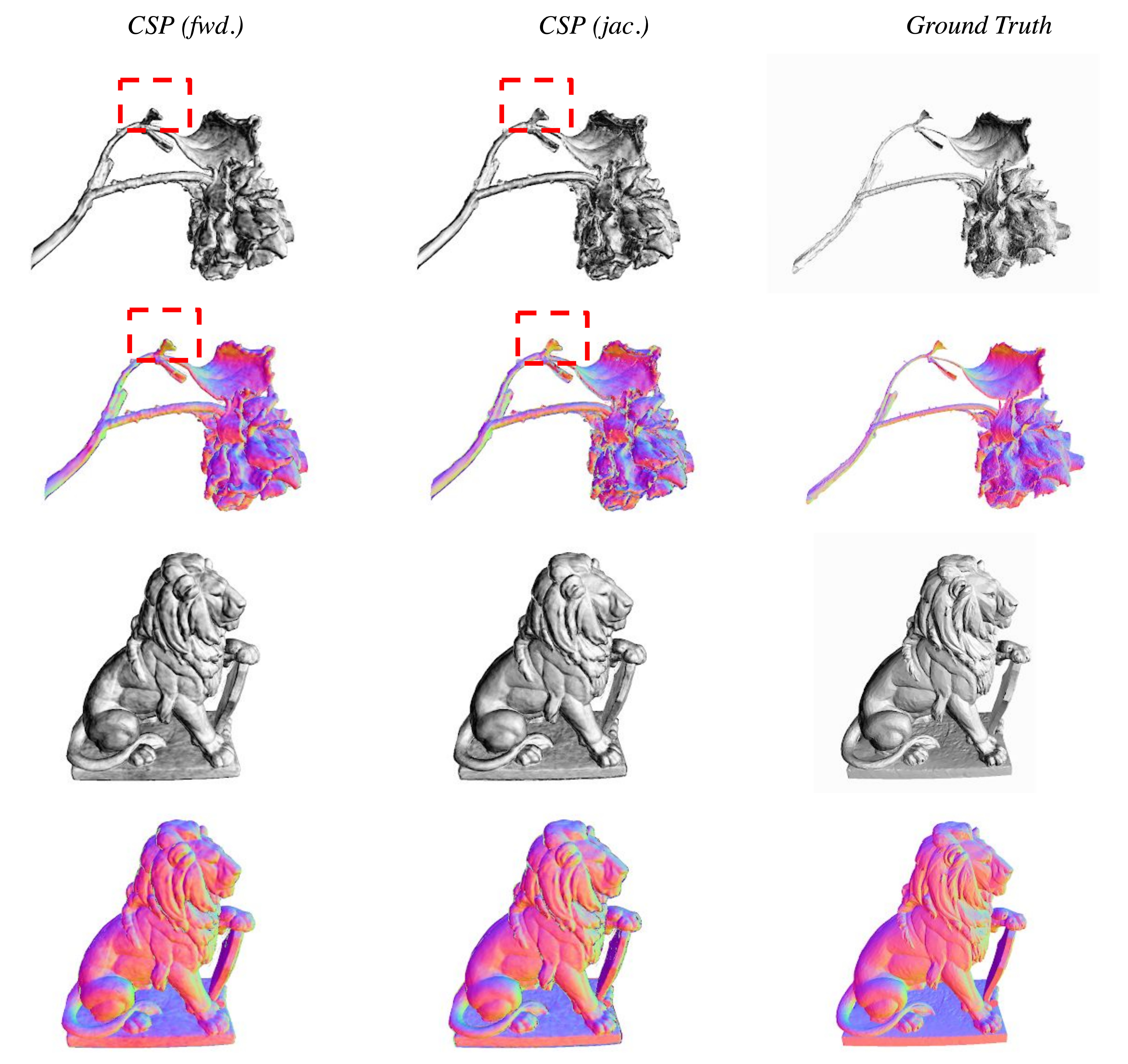}
    \caption{\textbf{Single Shape reconstructions: }Renderings from single shape architecture described in Sec.~\ref{sec:single_shape}. Here, we evaluate \name independently on two shapes with complex structures. We show lighted normals (row 1 of each shape) as well as the raw normal map (row 2 of each shape) using both normal estimation methods (see Sec.~\ref{sec:normals} in main paper) and compare against the ground truth for the same. The \name (jac.) results in higher quality normals compared to \name (\textit{fwd.}), which are reasonably comparable, but provide us with faster estimates (Highlighted in Red). More examples on next page.}
    \label{fig:ss_lighted_normals_1}
\end{figure*}
\begin{figure*}
    \centering
    \includegraphics[scale=0.20]{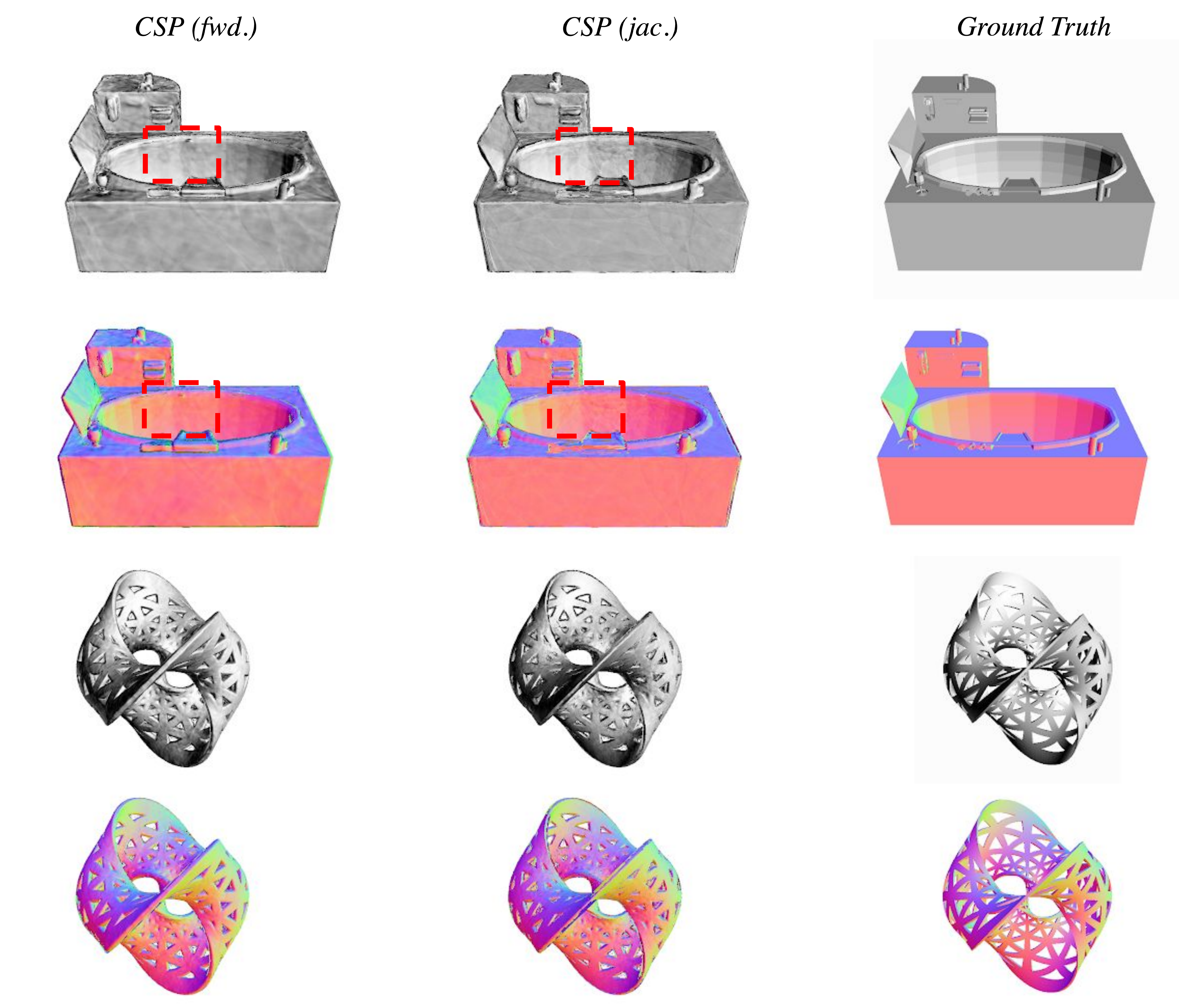}
    \caption{\textbf{Single Shape reconstructions: }Here, we show some results on a bathtub which has a high level of detail, with complex sub-structures, and a seifert surface which has complex topology (knots and holes).}
    \label{fig:ss_lighted_normals_2}
\end{figure*}

We present qualitative results for single shape reconstruction for a few complex shapes in Figures~\ref{fig:ss_lighted_normals_1} and ~\ref{fig:ss_lighted_normals_2}, illustrating the ability of \name to model complex shapes with high fidelity having either an open or a closed topology. It can be clearly seen that CSP is able to preserve surface details and accurately represent the surface orientations. In Figures~\ref{fig:ss_lighted_normals_1} and ~\ref{fig:ss_lighted_normals_2}, we present results on complex shapes like (a) a dried rose, and, (b) a lion statute having an intricate design and regions of varying curvature (c) a bathtub, that has high levels of detail and complex sub-structures, (d) the seifert surface~\cite{van2006visualization}, that has complex topology (multiple holes and knots), . 

\section{Training and architecture details}
\label{sec:train_arch}
This section shares the network architecture modeling the shape representation in Section~\ref{sec:shape space} of the main paper and details for training it. 

\par
We use the 3D volumetric encoder architecture proposed in ~\cite{10.1007/978-3-030-58580-8_31} with a \textit{\textbf{feature volume of resolution 64}}. Since our point estimation task is arguably more complex than binary occupancy prediction, we use a larger \textit{\textbf{decoder, with $512$ hidden units}} (with the same architecture as in ~\cite{10.1007/978-3-030-58580-8_31}). 

\par
We train with a \textit{\textbf{batch size of 32}} on different shapes, with an input point cloud of size 3000 (we follow the setup in NDF~\cite{NEURIPS2020_f69e505b}). For each shape in the batch, we use 10K points sampled from the training points, $\mathcal{P}$ (See Sec.~\ref{sec:data creat} of the main paper). We train on an NVIDIA GeForce RTX 2080Ti GPU using an ADAM~\cite{kingma2014adam} optimizer and a \textit{\textbf{learning rate of \num{1e-4}}}. It takes $\approx$ 5 days to train on the full ShapeNet dataset.

\section{Jacobian Computation: Implementation Details}
\label{sec:speed}
We now share the implementation details for the Jacobian computation as described in Sec.~\ref{sec:bwd_normals} of the main paper and discuss implications on its computational performance. 

The Jacobian is computed using 1 forward pass and 3 backward passes (one for each row of the Jacobian) through the same network. For this, we use the \codeword{autograd} package in PyTorch and set \codeword{retain_graph=True} when computing the first row of the Jacobian. This caches the activations in the graph and makes them readily available for computing the subsequent rows, speeding up the computation of the Jacobian.  

We logged the time taken to estimate the Jacobian matrix for the experiments described in Sec.~\ref{sec:rend_mesh} of the main paper for \textit{\name} (\textit{jac.}) and find that it takes on an average 0.08s for a 512$\times$512 image. In comparison, NDF is faster and takes 0.063s. This is to be expected as NDF just needs 1 forward and 1 backward pass. However, given that the computational graph needs to be obtained only once, we only incur an additional 25\% overhead (0.017s). Therefore, this is a reasonable trade-off for extracting high-fidelity surface normals. 

On the other hand, we also proposed an extremely fast method,  \textit{\name} (\textit{fwd.}),  which computes surface normals in a forward-mode taking only 0.003s for a 512$\times$512 image and is of a quality surpassing that of NDF (See Table~\ref{tab:normal_est} of main paper).

\section{Meshing and Rendering: Experimental setup}
\label{sec:rend_mesh_setup}
Results for setup used for rendering and meshing are presented in Section~\ref{sec:rend_mesh} of the main paper. Here we provide details of the experimental setup.

\noindent 
\textbf{Rendering.} For a given input point cloud, we first compute the 3D feature volume from the encoder. We then render the learnt \name representation (modeled using the decoder) from 3 different views. For doing so, we create a batch of rays from each viewpoint (3 views give us a total of 512$\times$512$\times$3$=$0.79M rays), and begin the sphere-tracing process (batched/parallel) for these set of rays. At the termination of sphere-tracing, we compute the surface normals for each ray (using gradients in case of NDF, and \textit{NVF} in case of \name). Since \textit{NVF} does not require a backward pass, it can accommodate a batch of 0.5M rays on a 8GiB GPU. The corresponding batch size for NDF is much lower at 0.15M since it requires the computation of gradients. As reported in Table \textcolor{red}{5} of the main paper, the increased batch size leads to a significant improvement in the rendering speed (i.e. time taken to compute the surface normals). \\  


\noindent 
\textbf{Meshing.} We present here additional details for meshing \textit{\name}\textit{s} using the novel coarse-to-fine meshing strategy outlined in Sec. {\color{red} 3.3.2} of the main paper. We compute a 3D distance grid (of resolution = $256$) using the proposed hierarchical space subdivision strategy, and perform meshing using Marching Cubes (using \textit{libmcubes}~\cite{stutz2018learning}) with a small positive threshold of $0.006$. For NDF, we use the code provided by authors to generate a dense point cloud (of 1M points) and mesh it using the ball-pivoting~\cite{bernardini1999ball} tool in meshlab~\cite{cignoni2008meshlab}, using a ball-radius of $0.01$. 

\par In our experiments, we have found the  ball-pivoting process to be very sensitive to this threshold, and in many cases it had to be tuned per-shape. On the other hand, our method uses a single threshold for all shapes, and generates high-fidelity meshes. Moreover, as reported in Sec.~\ref{sec:rend_mesh} of the main paper, our coarse-to-fine meshing strategy is significantly faster than that of NDF.  


\section{Additional qualitative results}
\label{sec:add_qual_res}
\begin{figure*}
    \centering
    \includegraphics[width=0.6\columnwidth]{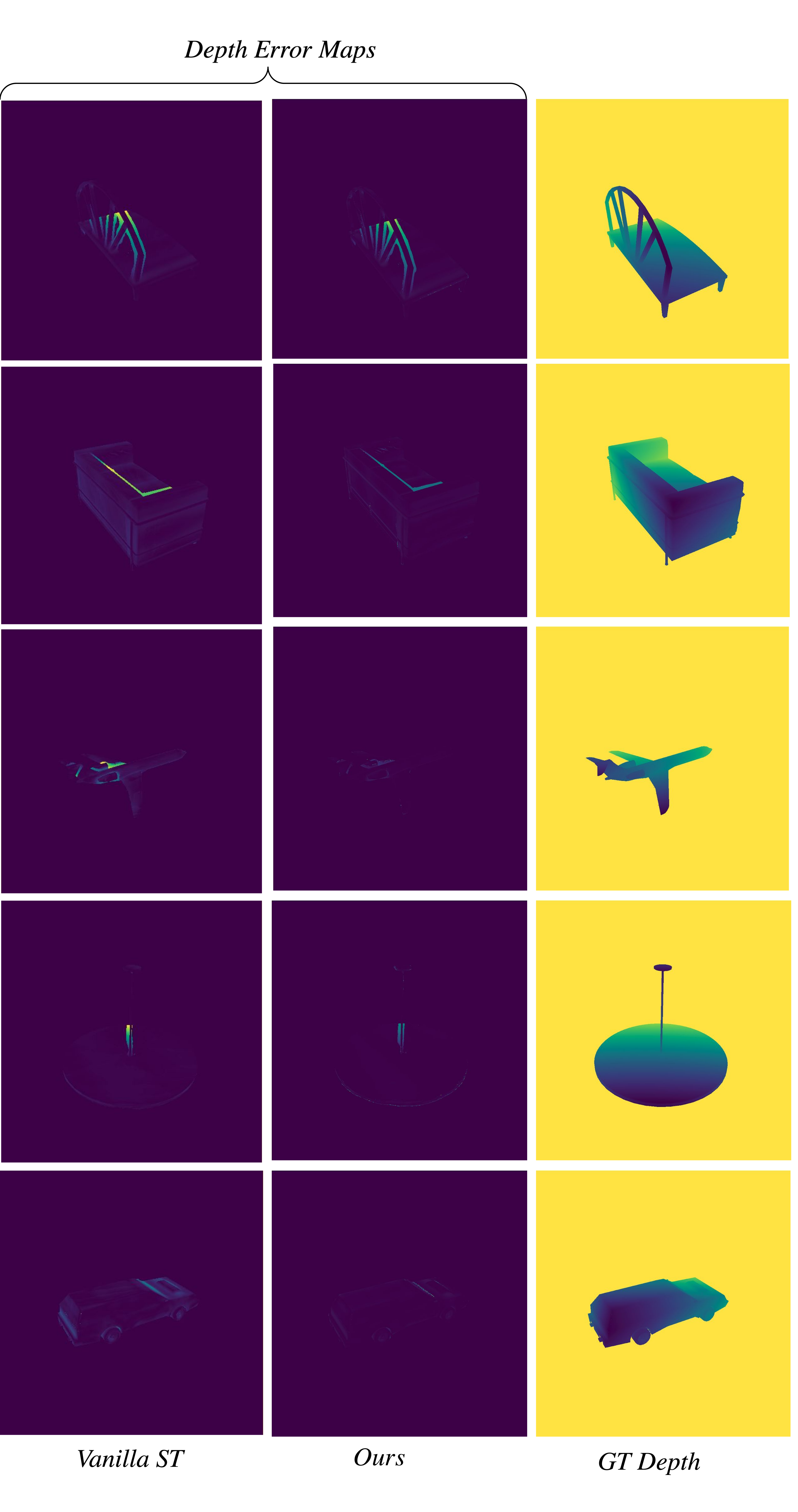}
    \caption{Comparison of depth maps generated by Vanilla Sphere Tracing (ST) and our novel projection-based algorithm outlined in Sec.~\ref{sec:spheretracing} of the main paper. We find that our method generates much lesser error when compared to the conventional sphere-tracing strategy.}
    \label{fig:sup_depth_maps}
\end{figure*}

\begin{figure*}
    \centering
    \includegraphics[width=0.6\columnwidth]{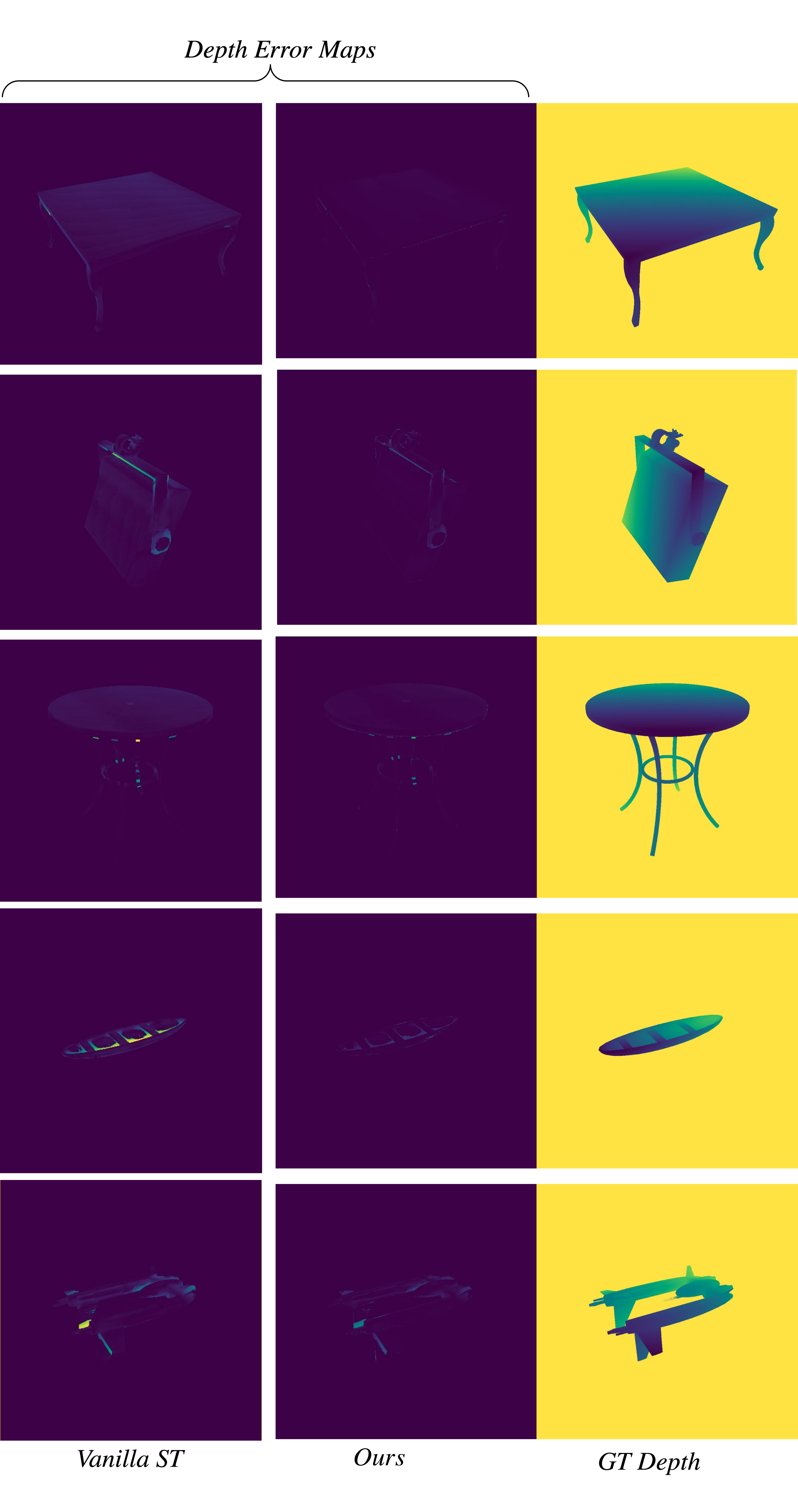}
    \caption{Additional results showing depth error maps.}
    \label{fig:sup_depth_maps_1}
\end{figure*}

\begin{figure*}
    \centering
    \includegraphics[width=0.9\columnwidth]{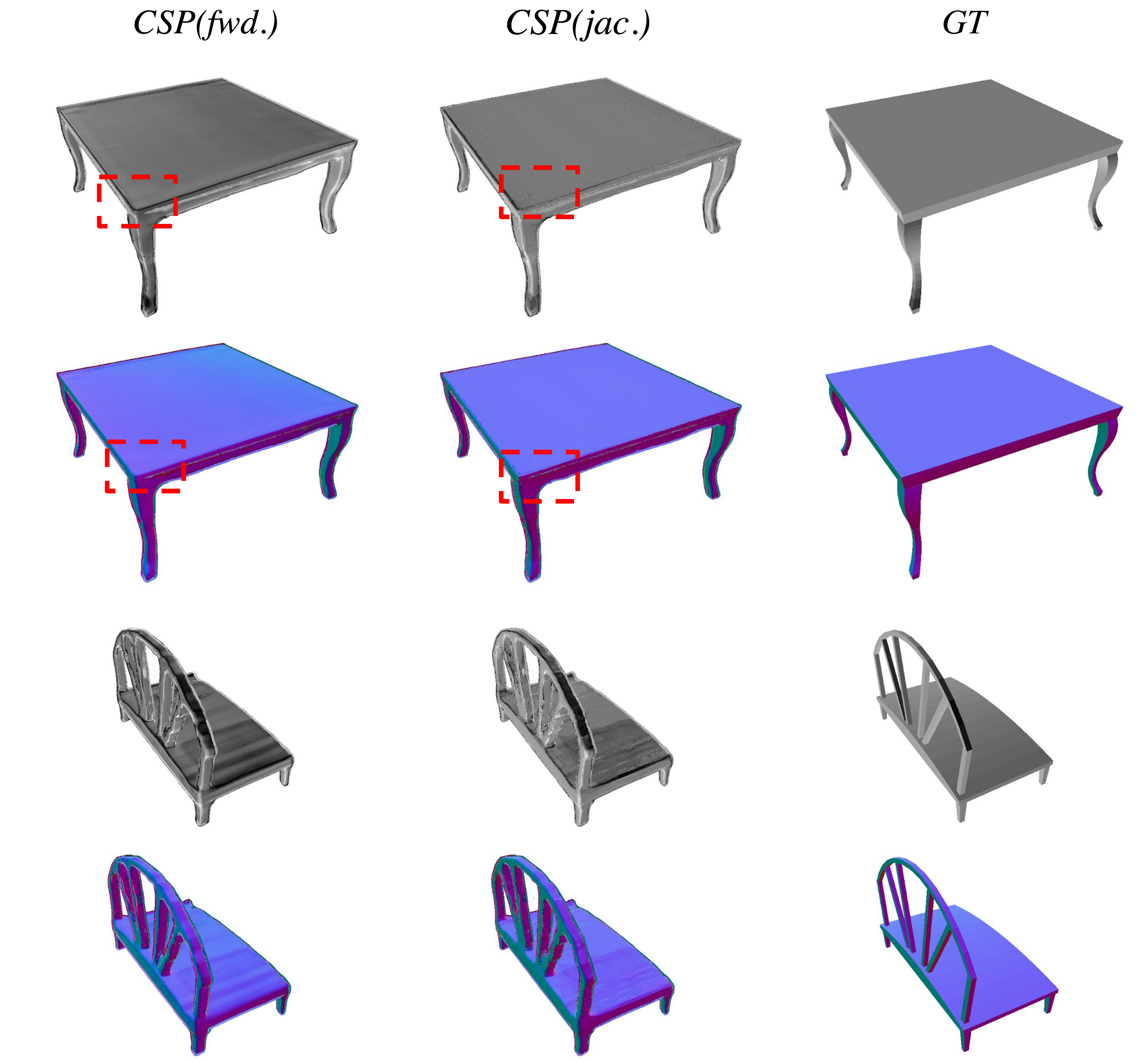}
    \caption{Surface reconstruction results on exemplar shapes from ShapeNet test set. Here, we show both \name (\textit{jac.}) and \name (\textit{fwd.}) ($\alpha=0.005$) side-by-side, with the first row of each shape depicting a rendering of the sphere-traced surface normal map (shown in the second row) with directional light. We find that both methods (see Sec.~\ref{sec:normals} for a description of these methods, and Sec.~\ref{sec:local surface prop} for some initial results reported in main paper) yield high-quality surface normals (with \textit{\name } \textit{(fwd.)}) providing efficient forward-mode normal estimates. Note also that \textit{\name } \textit{(jac.)} is marginally better in some regions (Highlighted in red.).}
    \label{fig:sup_main_fig}
\end{figure*}

\begin{figure*}
    \centering
    \includegraphics[width=0.9\columnwidth]{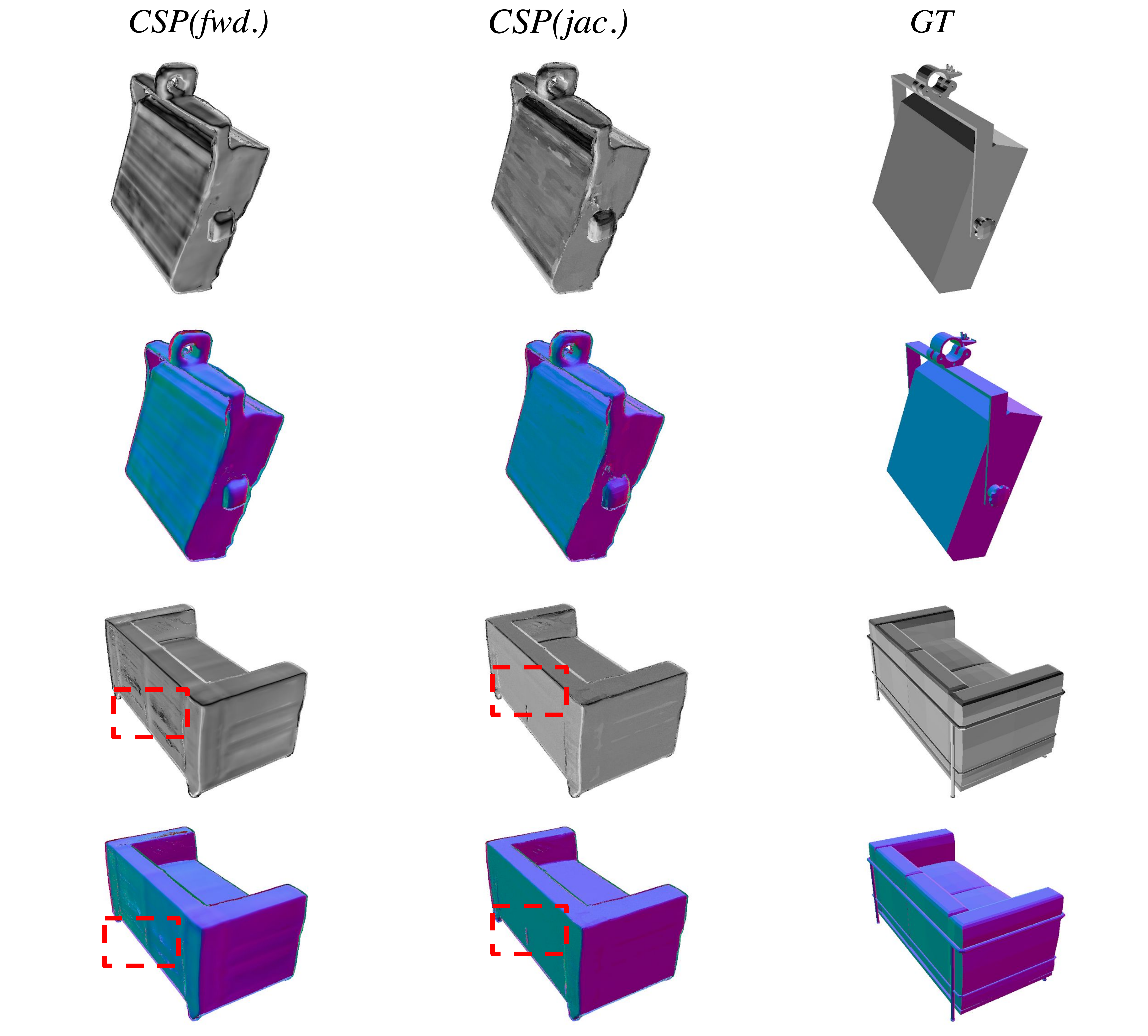}
    \caption{Additional surface reconstruction results from ShapeNet test set.}
    \label{fig:sup_main_fig_1}
\end{figure*}

\begin{figure*}
    \centering
    \includegraphics[width=0.9\columnwidth]{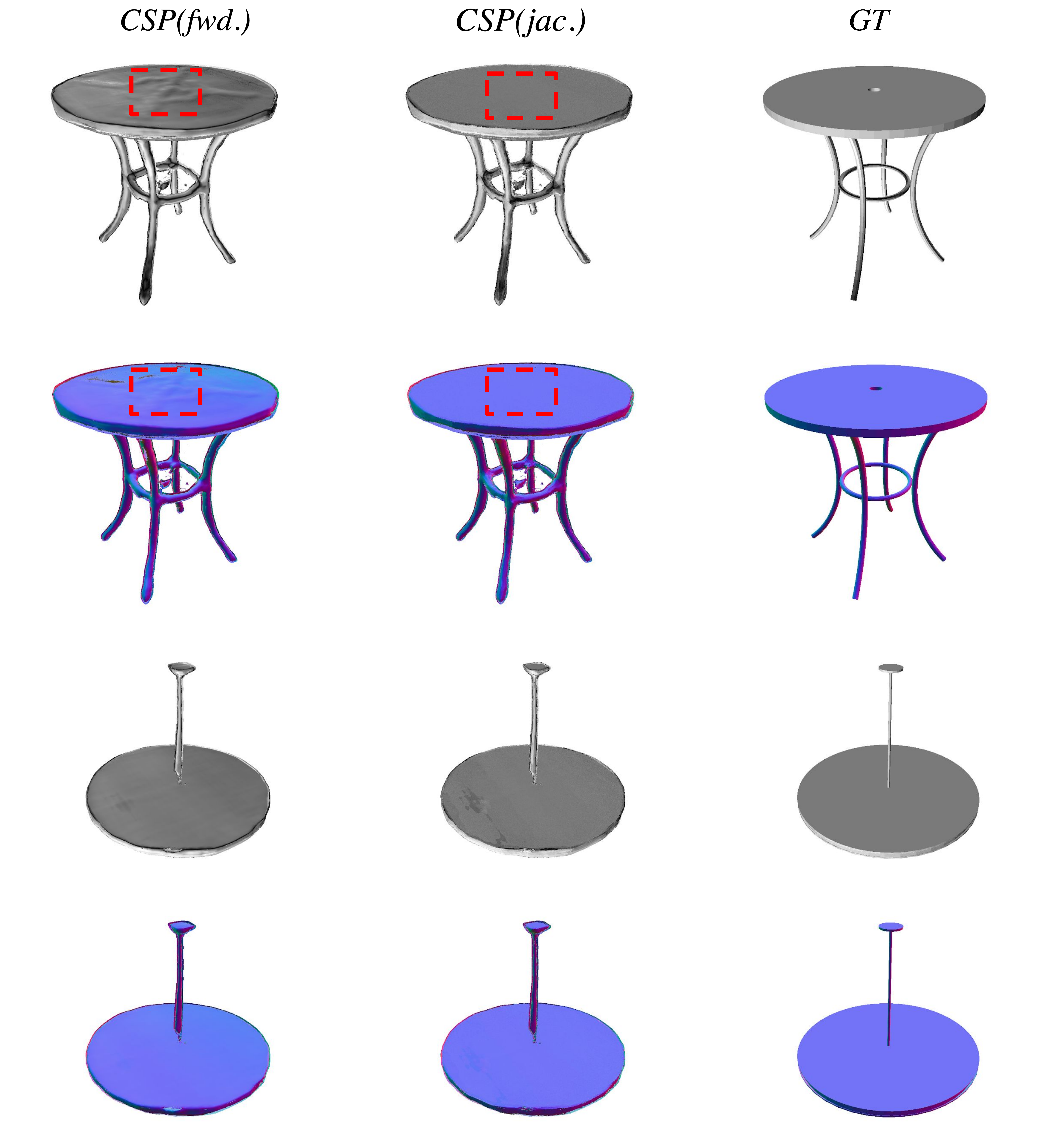}
    \caption{Additional surface reconstruction results from ShapeNet test set.}
    \label{fig:sup_main_fig_2}
\end{figure*}

\begin{figure*}
    \centering
    \includegraphics[width=0.9\columnwidth]{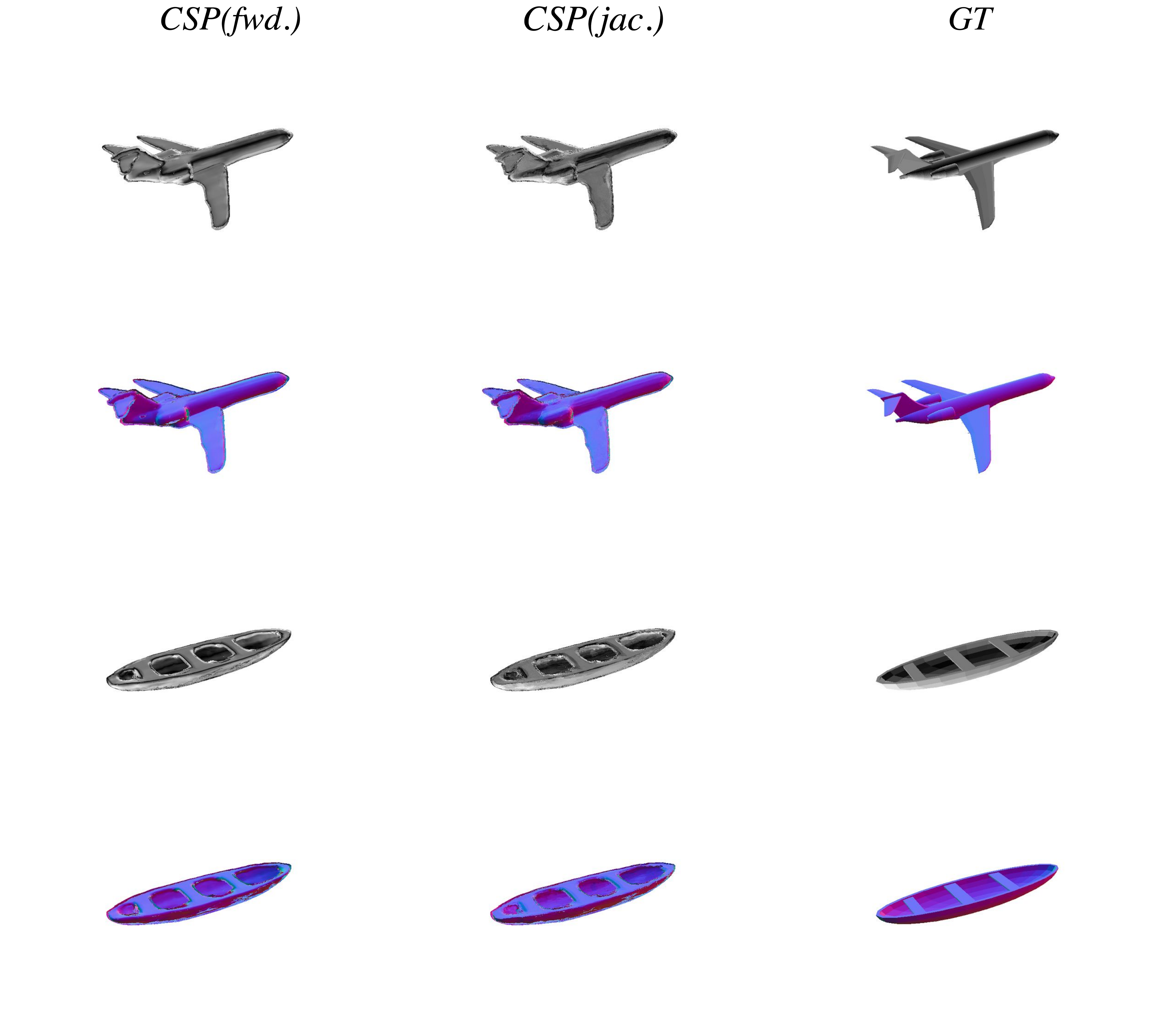}
    \caption{Additional surface reconstruction results from ShapeNet test set.}
    \label{fig:sup_main_fig_3}
\end{figure*}

\begin{figure*}
    \centering
    \includegraphics[width=0.9\columnwidth]{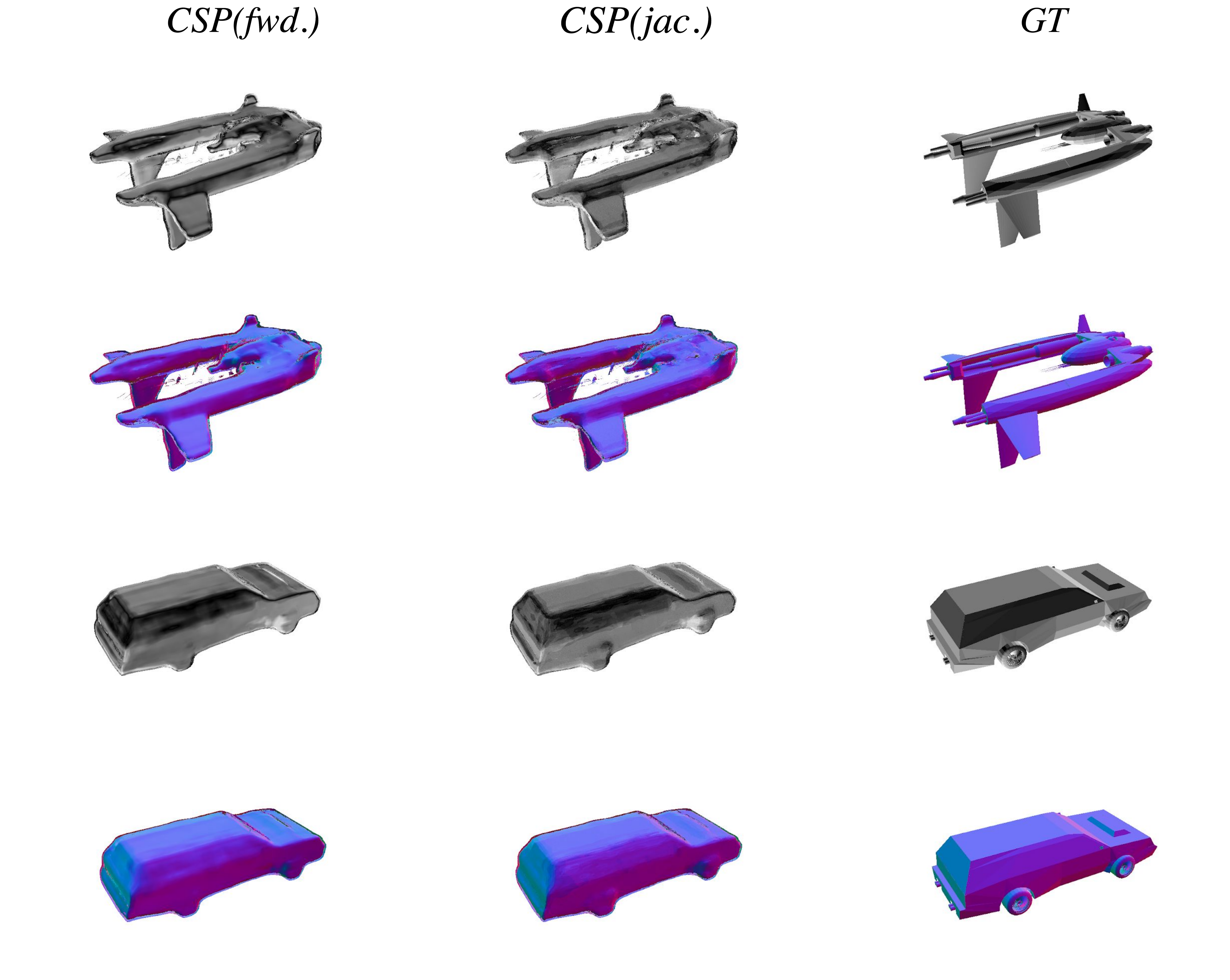}
    \caption{Additional surface reconstruction results from ShapeNet test set.}
    \label{fig:sup_main_fig_4}
\end{figure*}

\begin{figure*}
    \centering
    \includegraphics[width=0.9\columnwidth]{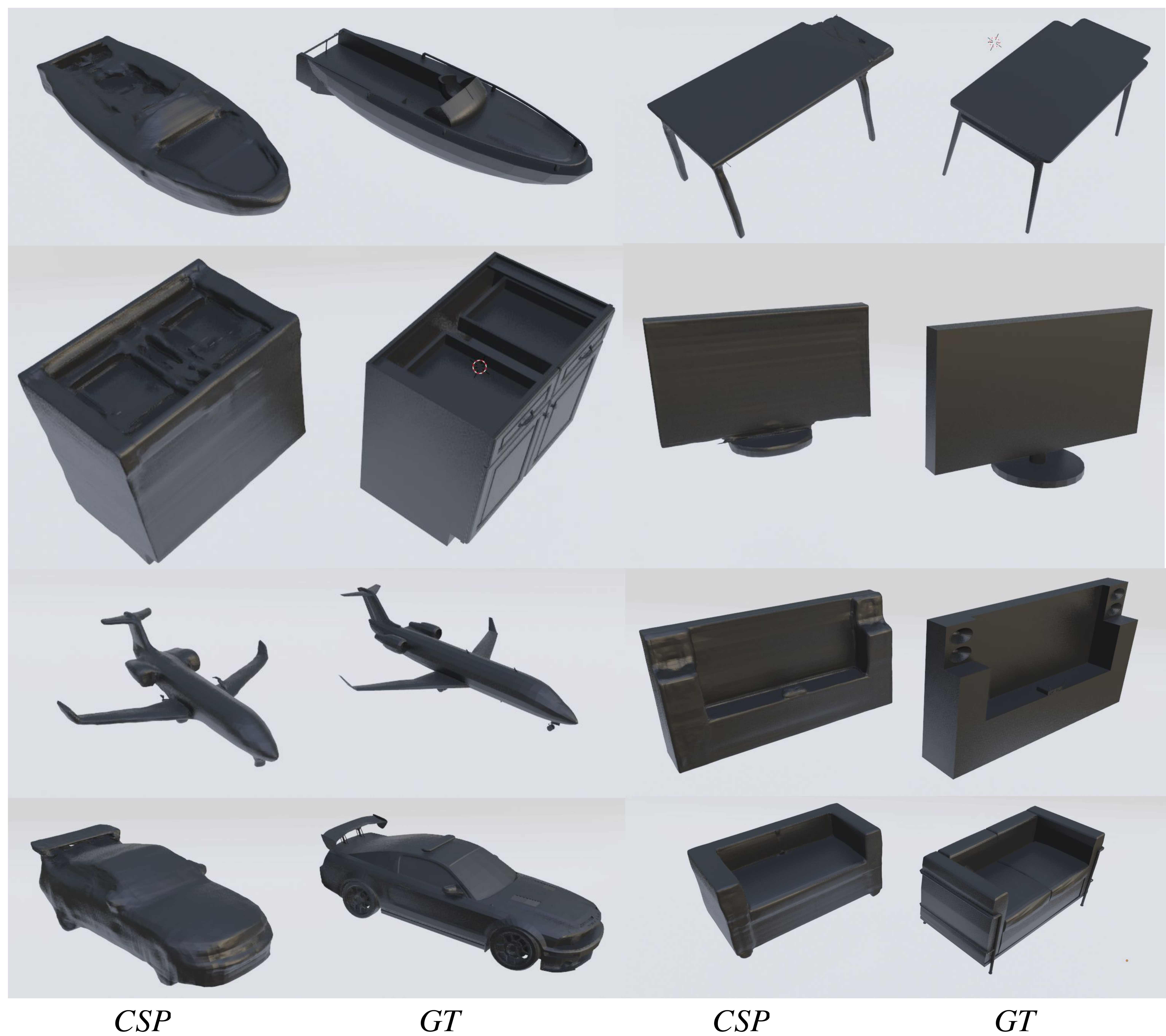}
    \caption{Meshes generated by our novel coarse-to-fine meshing algorithm for \textit{\name}\textit{s} (see Sec. \textcolor{red}{3.3.2} of main paper). We also show the Ground Truth mesh on the right of each subfigure. Note that our algorithm generates structurally consistent meshes, which render visually pleasing images in Blender~\cite{blender}.}
    \label{fig:sup_meshes}
\end{figure*}

To supplement the qualitative results on the various sphere-tracing strategies (Fig.~\ref{fig:depth_error} of main paper), in Fig.~\ref{fig:sup_depth_maps}, ~\ref{fig:sup_depth_maps_1}, we show additional results which compare \textbf{\textit{depth maps}} generated using our novel sphere-tracing algorithm for \name, against a vanilla sphere-tracing technique for unsigned distance functions. Further, in Fig.~\ref{fig:sup_main_fig}, ~\ref{fig:sup_main_fig_1}, ~\ref{fig:sup_main_fig_2}, ~\ref{fig:sup_main_fig_3}, ~\ref{fig:sup_main_fig_4} we show additional examples of shape reconstruction which bolster the results shown in Fig.~\ref{fig:shape_recon} of the main paper, and demonstrate the capability of our class-agnostic model to reconstruct shapes from any class of ShapeNet. All results are shown on a test-set of shapes (ShapeNet test-set used in~\cite{psgn}) not seen in training. Additionally, to reiterate the utility of meshes generated by our novel meshing algorithm for \textit{\name}\textit{s} (Sec.~\ref{sec:meshing} of main paper), we also show some \textit{\textbf{representative meshes}} (compared against GT meshes) generated in   Fig.~\ref{fig:sup_meshes}.

\section{Off The Shelf Tools and Packages Used}
\label{sec:tools}
In this work, we make use of a variety of off-the-shelf packages to run our experiments. For generating data, we use \codeword{faiss}~\cite{JDH17}, which is a library for performing fast nearest neighbour search on GPU. We compute GT normal and depth maps using the \codeword{trimesh}~\cite{trimesh} with \codeword{pyembree} bindings viz. \codeword{trimesh.ray.ray_pyembree.RayMeshIntersector}. \codeword{torch-scatter}\footnote{\href{https://github.com/rusty1s/pytorch\_scatter}{https://github.com/rusty1s/pytorch\_scatter}} is used for trilinear interpolation of the 3D Feature Volume (See Fig.~\ref{fig:sys_diag} of the main paper). For sphere-tracing \name, we provision a custom implementation in \codeword{PyTorch}, which renders multiple images efficiently by batching rays across different views.



\end{document}